\documentclass{article}

\PassOptionsToPackage{authoryear, sort&compress}{natbib}

\usepackage[final]{neurips_2024}

\usepackage[colorinlistoftodos,prependcaption,textsize=tiny]{todonotes}
\usepackage{xargs} 
\usepackage{wrapfig}
\usepackage{threeparttable}
\usepackage{enumitem}
\usepackage{makecell}
\usepackage[utf8]{inputenc} 
\usepackage[T1]{fontenc}    
\usepackage{url}            
\usepackage{booktabs}       
\usepackage{amsfonts}       
\usepackage{nicefrac}       
\usepackage{xcolor}         
\usepackage{microtype}
\usepackage{graphicx}
\usepackage{floatrow}
\usepackage{subcaption}
\usepackage{listings}
\usepackage{longtable}
\usepackage{array}
\usepackage{tcolorbox} 
\usepackage{multirow}
\usepackage{marvosym} 

\usepackage[bookmarks=true,colorlinks=true,linkcolor=black,urlcolor=magenta,citecolor=green!40!black]{hyperref}

\usepackage{amsmath}
\usepackage{amssymb}
\usepackage{mathtools}
\usepackage{amsthm}
\theoremstyle{plain}

\theoremstyle{definition}

\theoremstyle{remark}

\usepackage{bm} 
\usepackage{algorithm, algorithmic}

\usepackage{etoolbox}
\usepackage{natbib}
\newcounter{bibcount}
\makeatletter
\patchcmd{\@lbibitem}{\item[}{\item[\hfil\hspace{2.5em}\stepcounter{bibcount}{[\thebibcount]}\hspace{0.em}}{}{}
\makeatother
\usepackage[nameinlink,capitalise]{cleveref}
\crefname{section}{§}{§§}
\Crefname{section}{§}{§§}

\usepackage{pifont}

\newcommand{\our}{LLM Psychometrics}

\title{
    Large language model psychometrics:\\A systematic review of\\evaluation, validation, and enhancement
}

\author{%
  Haoran Ye\textsuperscript{1}, Jing Jin\textsuperscript{1}, Yuhang Xie\textsuperscript{1}, Xin Zhang\textsuperscript{2,3}, Guojie Song \textsuperscript{1,\Letter}
  \\[0.1cm]
\textsuperscript{1}State Key Laboratory of General Artificial Intelligence,\\School of Intelligence Science and Technology, Peking University\\
\textsuperscript{2}School of Psychological and Cognitive Sciences, Peking University\\
\textsuperscript{3}Key Laboratory of Machine Perception (Ministry of Education), Peking University\\[0.1cm]
\small \texttt{hrye@stu.pku.edu.cn} \quad \texttt{gjsong@pku.edu.cn}\\[0.5cm]
Project Website: \href{https://llm-psychometrics.com}{https://llm-psychometrics.com}\\
}

\begin{document}

\maketitle

\begin{figure}[h]
    \centering
    \includegraphics[width=0.3\textwidth]{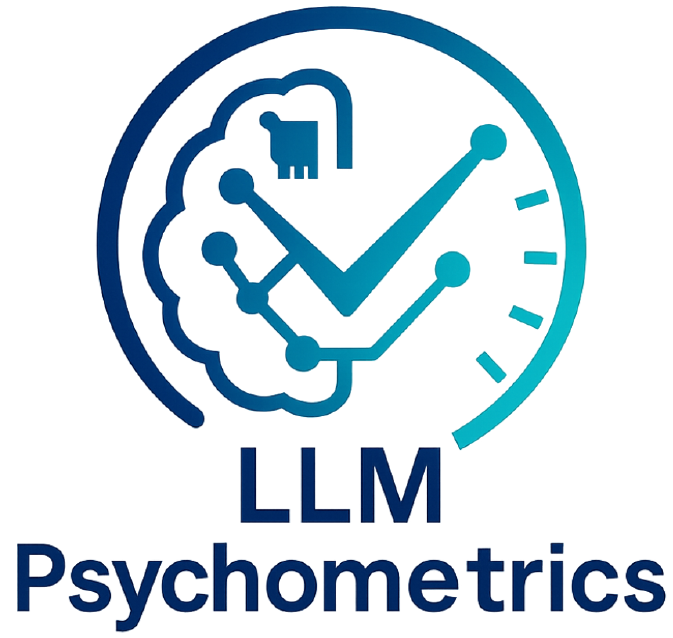}
\end{figure}
\vspace{0.4cm}

\begin{abstract}
The advancement of large language models (LLMs) has outpaced traditional evaluation methodologies. This progress presents novel challenges, such as measuring human-like psychological constructs, moving beyond static and task-specific benchmarks, and establishing human-centered evaluation. These challenges intersect with psychometrics, the science of quantifying the intangible aspects of human psychology, such as personality, values, and intelligence. This review paper introduces and synthesizes the emerging interdisciplinary field of \our{}, which leverages psychometric instruments, theories, and principles to evaluate, understand, and enhance LLMs. The reviewed literature systematically shapes benchmarking principles, broadens evaluation scopes, refines methodologies, validates results, and advances LLM capabilities. Diverse perspectives are integrated to provide a structured framework for researchers across disciplines, enabling a more comprehensive understanding of this nascent field. Ultimately, the review provides actionable insights for developing future evaluation paradigms that align with human-level AI and promote the advancement of human-centered AI systems for societal benefit.
A curated repository of LLM psychometric resources is available at \href{https://github.com/valuebyte-ai/Awesome-LLM-Psychometrics}{https://github.com/valuebyte-ai/Awesome-LLM-Psychometrics}.
\end{abstract}

\clearpage

{\color{black}
\tableofcontents
}

\hypersetup{linkcolor=red}

\clearpage

\section{Introduction}\label{sec: introduction}

The advent of large language models (LLMs) represents a transformative breakthrough in AI. These systems exhibit general-purpose capabilities spanning diverse domains \citep{bubeck2023sparks}, with particular proficiency in natural language understanding and generation \citep{demszky2023using, grossmann2023ai, ziems2024can, gu2024llm-as-a-judge}. They are rapidly integrated into critical societal infrastructure, ranging from consumer-facing applications like chatbots \citep{chatgpt} and search engines \citep{wang2024large_search_model} to high-stakes domains such as healthcare \citep{singhal2023large}, education \citep{milano2023large}, and scientific discovery \citep{romera2024mathematical,ye2024reevo}.
Their increasing dominance has exposed a fundamental, pressing scientific challenge: how can we rigorously evaluate these AI systems that transcend traditional benchmarks of biological or algorithmic intelligence? 

Traditional AI evaluation has relied on curating task-specific datasets, annotating ground-truth labels with human input, running models on these datasets, and assessing performance using predefined metrics. 
However, LLMs have triggered an evaluation crisis, as their versatile capabilities and human-like behaviors exceed what traditional benchmarks can measure.
Novel challenges include but are not limited to (1) evaluating psychological constructs like personality, values, and cognitive biases; (2) obsolescence of static benchmarks due to rapid LLM development and training data contamination; (3) compromised robustness and validity due to LLMs' prompt- and context-sensitivity; (4) requiring human-centered evaluation approaches; and (5) expanding evaluation methodologies in scope and complexity as LLMs integrate into agentic and multimodal systems.

These challenges intersect with humanity's century-old quest to quantify the complex, intangible human psychology \citep{pasquali2009psychometrics}. Psychometrics emerged from this timeless pursuit as the scientific study of psychological measurement. It bridges the abstract and the empirical by transforming human traits into quantifiable data, enabling better understanding, prediction, and decision-making in education, business, healthcare, governance, and beyond \citep{rust2014modern_psychometrics}.

This intersection sets the stage for a new research frontier that we term \textbf{\our{}}. This interdisciplinary field is dedicated to evaluating, understanding, and enhancing LLMs through the application and integration of psychometric instruments, theories, and principles.
Crucially, \our{} treats LLMs themselves as the measured subjects, rather than instruments for measuring humans. The field seeks to quantify, interpret, manipulate, and improve the human-like attributes and behaviors exhibited by LLMs, spanning non-cognitive constructs (e.g., personality, values, morality, attitudes) and cognitive constructs (e.g., heuristics and biases, social interaction abilities, psycholinguistic abilities, learning and reasoning capacities). Grounded in psychometric principles, research in \our{} examines measurement properties (e.g., reliability, validity, generalizability, and measurement invariance) and uses these insights to inform targeted model improvement.

Recent research in \our{} pioneers in addressing the LLM evaluation crisis. Some studies introduce dynamic and construct-oriented evaluation frameworks that move beyond static, task-specific benchmarks \citep{zhu2024dynamic,hagendorff2023machine}.
In parallel, novel methodologies are developed to measure non-cognitive constructs \citep{ren2024valuebench, huang2023psychobench, pellert2024ai_psychometrics}. Self-adaptive evaluation techniques now allow for the extrapolation of item difficulty and tailoring assessments to model performance \citep{jiang2024raising, lalor2024item, polo2024tinybenchmarks}. Drawing from the methodological framework of psychometric validation, research improves the reliability and validity of evaluation protocols \citep{ye2025gpv}. Human-centered evaluation drives aligning model behavior with human values \citep{wang2024incharacter, yao2025value_compass_leaderboard}. In addition, the scope of evaluation expands to agentic and multimodal systems, further broadening the methodological landscape \citep{li2024quantifying,huang2024designing}.

Notably, we view \our{} as a methodological borrowing focused on behavioral manifestation. We do not posit that LLMs possess subjective experience, sentience, or developmental histories. Rather, our focus is on how psychological constructs manifest in their outputs, as well as the fidelity, consistency, and controllability of these behavioral manifestations. 
Accordingly, when we refer to a \textit{construct} in the context of LLMs throughout this review, we mean a \textit{synthetic behavioral manifestation}: systematic response patterns measurable via psychometric frameworks, rather than a claim about an underlying genuine psychological state.
Given the increasing integration of LLMs into human-facing roles, measuring and controlling these manifested behaviors holds immense practical value for user experience, safety, and alignment, regardless of the models' internal ontological status. However, realizing this practical value is non-trivial. The very act of applying anthropocentric tests to non-human entities raises unresolved questions about reliability and validity. A fundamental debate continues regarding whether these instruments capture meaningful latent patterns in LLMs or merely reflect sophisticated statistical mimicry. Furthermore, the sensitivity of LLMs to prompts, decoding strategies, and context strains core psychometric assumptions such as trait stability.

Amidst the debates, the field of \our{} has seen significant growth, evidenced by the proliferation of related research papers.
These studies, however, often operate in silos, addressing isolated psychological constructs, employing diverse methodologies, and utilizing distinct validation techniques.
The interdisciplinary nature of this domain has attracted contributions from a broad spectrum of academic communities, yet there remains a lack of cohesion among researchers. This fragmentation of insights, particularly between studies focusing on disparate constructs, underscores a pressing need for a systematic review to synthesize these efforts and facilitate a more integrated, scientifically robust understanding of the field.

\begin{figure}[!t]
    \centering
    \includegraphics[width=\textwidth]{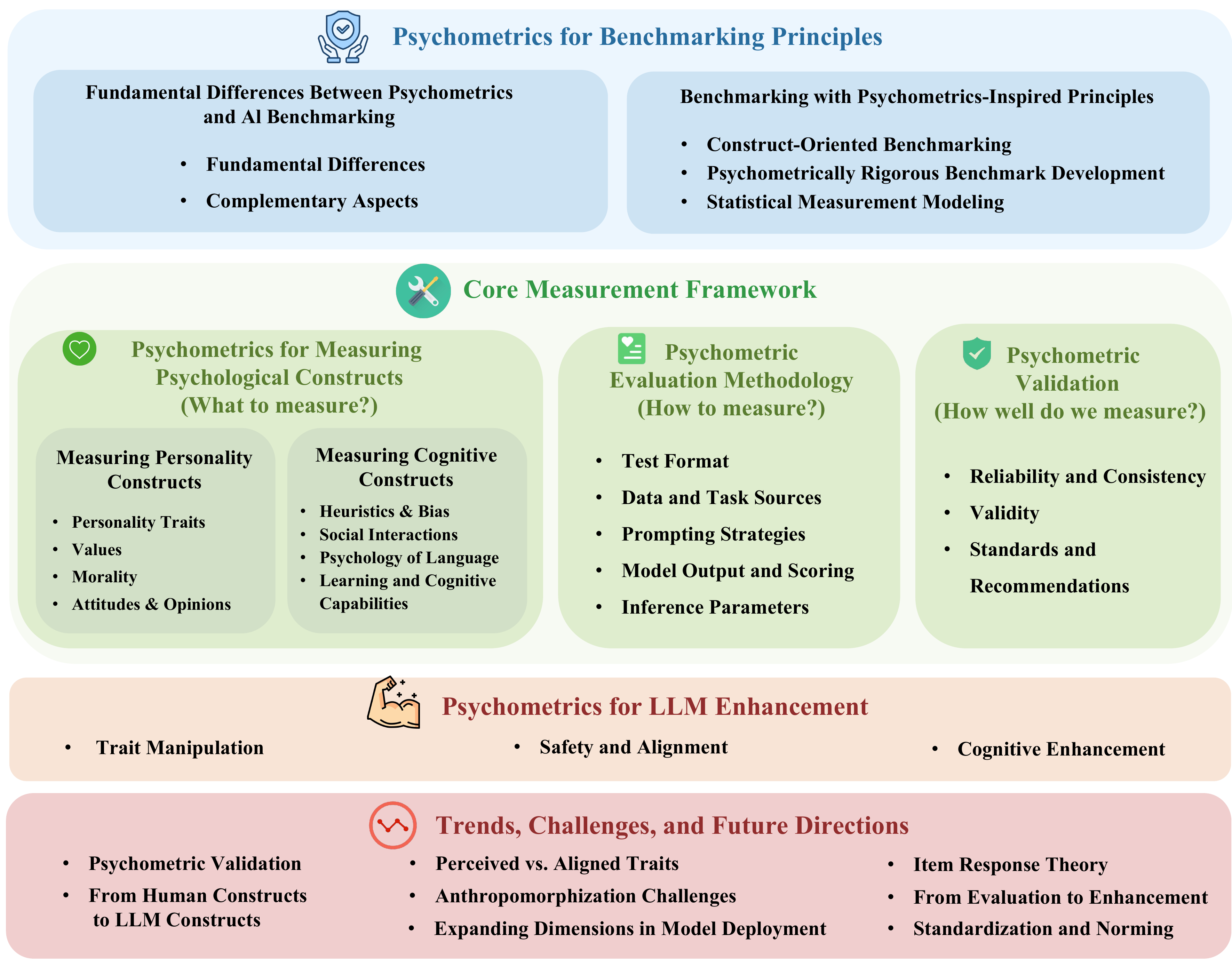}
    \caption{Overview of this review.}
    \label{fig: overview}
\end{figure}

\paragraph{Taxonomy and paper structure.}
To bridge the gap, we systematically review \our{} across evaluation, validation, and enhancement. 
The evaluation framework encompasses three core dimensions: the target construct (\textit{what to measure}), the measurement method (\textit{how to measure}), and the validation of results (\textit{how well do we measure}). The psychometric insights not only inform evaluation but also guide the development and refinement of LLMs (\textit{how to improve}).
Accordingly, this review is structured as follows (\cref{fig: overview}).
\cref{sec: preliminaries} provides an overview of the preliminaries and methodological foundations to facilitate subsequent discussions.
\cref{sec: principles} thoroughly contrasts psychometrics and traditional AI benchmarking and reviews how psychometric principles can underpin and reinvent LLM benchmarks.
The core measurement framework is detailed in \cref{sec: construct}, \cref{sec: method}, and \cref{sec: validation}.
\cref{sec: construct} delves into the psychological constructs evaluated in LLMs, elucidating the theories employed and summarizing key evaluation findings. In \cref{sec: method}, we scrutinize the psychometric evaluation methodologies applied to LLMs, followed by \cref{sec: validation}, which examines the psychometric validation of evaluation results. Beyond evaluation, \cref{sec: enhancement} introduces strategies for enhancing LLMs through psychometric insights. \cref{sec: trends} discusses emerging trends, challenges, and future directions. Finally,
\cref{sec: ethics} highlights key ethical considerations, while \cref{sec: conclusion} concludes the paper.

\paragraph{Scope and related work.}
This review treats LLMs as subjects of psychometric tests. We distinguish this from using LLMs as tools to assess or profile human respondents, which is out of scope; see \citet{ye2025psychometrics} for a review of that adjacent field. We also exclude evaluation studies if they do not employ psychometric approaches, adhere to psychometric principles, or if they focus solely on scalar performance metrics rather than characterizing latent behavioral tendencies.
The operational distinction between psychometric research and conventional AI benchmarking, which underpins these inclusion criteria, is elaborated in \cref{sec: psy_bench_diff}.
Studies that merely administer scales and report scores meet the minimal inclusion threshold but offer limited scientific value. Rigorous \our{} additionally evaluates measurement properties, such as reliability across prompt variants and construct validity via factor analysis.
Readers interested in conventional LLM benchmarking are encouraged to consult surveys on LLM evaluation \citep{guo2023survey_llm_eval, chang2024survey_llm_eval}. Several related reviews focus on the evaluation of specific constructs in LLMs, such as personality \citep{wen2024llm_personality_survey,dong2025humanizing}, attitudes and values \citep{ma2024aov_survey}, cultural awareness \citep{pawar2024culture_survey,adilazuarda2024culture_survey}, and theory of mind \citep{saritacs2025tom_survey,dong2025humanizing}.
In addition, fairness and related psychometric concepts (e.g., differential item functioning, measurement invariance) are primarily discussed in the context of using LLMs as tools for human psychometrics.
We refer readers interested in AI fairness to dedicated surveys \citep{mehrabi2021survey,gallegos2024bias_fairness_survey}.
\citet{hagendorff2023machine,hagendorff2024machine} introduce the notion of machine psychology and discuss the emergent LLM abilities; however, they do not provide comprehensive coverage of the related research, nor do they elaborate on LLM personality constructs, psychometric validation, or enhancement. This paper provides the first systematic review of \our{}.

\section{Preliminary and methodological foundations}\label{sec: preliminaries}

We aim for our review to be self-contained and accessible to a broad, cross-disciplinary audience. To this end, this section presents the preliminary and methodological foundations that underpin the subsequent discussions.

\subsection{Large language models}\label{sec: llms}
LLMs are large-scale \textit{deep neural networks}—essentially complex systems of nonlinear regression equations.
An LLM can generate text by predicting the next \textit{token} (word or subword) in a sequential manner (autoregressive generation), given the preceding context. It does so by modelling a conditional probability distribution over the vocabulary; i.e., the likelihood of each token given the context:
\begin{equation}
    P(x_t | x_{<t}) = f(x_{<t}; \theta),
    \label{eq: llm_conditional_probability}
\end{equation}
where $x_t$ is the token at time step $t$; $x_{<t}$ is the context preceding $x_t$, usually including both user prompts and previously generated tokens; $f$ is the model's parameterized function; and $\theta$ represents the model parameters. Given $f$, the model generates text by either sampling from this distribution or directly selecting the token with the highest probability.
In the former case, hyperparameters such as \textit{temperature} can be adjusted to control the diversity of generated text. In the latter case, the model is said to use \textit{greedy decoding}, and the generated text is deterministic.
In evaluating LLMs, it is crucial to properly account for the stochasticity of the model.

These models are based primarily on the \textit{transformer} architecture, a neural network design that employs self-attention mechanisms to capture contextual relationships between words, phrases, and broader linguistic patterns. Modern LLMs typically contain billions of parameters, enabling them to efficiently learn from vast amounts of textual data.
During evaluation, if the model has already been exposed to the test items during training, this is referred to as \textit{data contamination}. In such cases, the model is more likely to exhibit artificially inflated performance or simply reproduce memorized patterns, rather than revealing its true underlying capabilities or traits.

The training process of LLMs is typically divided into two phases: \textit{pre-training} and \textit{post-training}.
\textit{Pre-training} is the phase where LLMs learn to predict the next token, given its preceding context, on a large corpus of text data. This process is unsupervised, as the model does not require explicit labels or annotations to learn the underlying patterns in the data. The model processes Internet-scale text data from diverse sources like books, articles, and websites. By repeatedly predicting the next word in a sentence, the model learns the statistical properties of language and gains large-scale world knowledge.
Models that have only undergone the pre-training phase are usually referred to as \textit{base models}.
\textit{Post-training}, or \textit{fine-tuning}, is the process of adapting the base models to better follow user instructions, align with human values, or specialize in particular tasks. 
This stage typically involves training the model on a smaller, human-annotated dataset or incorporating human feedback on the quality of the model outputs. Models that have undergone both phases are often referred to as \textit{fine-tuned models}, \textit{instruction-tuned models}, or \textit{aligned models}.

We interact with LLMs using \textit{prompts}, which are input instructions to the model. For psychometric evaluation, these prompts can naively be the reformatted versions of test items originally designed for humans, adapted for LLMs to answer. When designing prompts, one should consider differences between base and fine-tuned models. Most public-facing LLMs are fine-tuned, so evaluation research primarily focuses on these models for greater practical relevance.

A key emergent capability of LLMs is \textit{in-context learning}, which allows models to adapt to new tasks or patterns by conditioning on examples or instructions provided within the input context \(x_{<t}\), without modifying model parameters. This property can influence LLM performance in psychometric evaluation. For instance, prompting models to reason step-by-step (e.g., Chain-of-Thought prompting \citep{wei2022chain}) can enhance performance on reasoning tasks, while instructing them to role-play may modulate their exhibited personalities and values.

\subsection{Psychometrics}\label{sec: psychometrics}

Psychometrics, also known as psychological testing, involves the use of tests to measure, understand, or predict behavior by quantifying specific actions or characteristics. These tests rely on samples of behavior, meaning they are not perfect measures and often include errors inherent to sampling. Test \textit{items} are specific stimuli designed to elicit observable reactions that can be scored or evaluated. Typically, tests are composed of multiple questions or problems as their items, producing explicit data subject to scientific analysis.

A \textit{psychological test} is a set of items that are designed to measure characteristics of human beings that pertain to behavior \citep{kaplan2001psychological_testing_book}. 
Behavior measured by tests can be \textit{overt} (observable actions) or \textit{covert} (internal thoughts or feelings). Tests may assess past, current, or even predict future behavior. Interpretation of test scores depends on their context within a distribution. \textit{Scales} are used to relate raw scores to defined distributions, aiding interpretation. In addition, psychological tests can measure \textit{traits}—enduring tendencies like shyness or determination—and \textit{states}, which reflect temporary conditions of individuals.

Psychological testing measures \textit{individual differences} in various \textit{constructs}, which are abstract psychological attributes or dimensions that help explain and predict behavior. Two primary categories of such constructs are \textit{personality constructs} and \textit{cognitive constructs} \citep{kaplan2001psychological_testing_book}. \textit{Personality tests} focus on an individual's tendencies and dispositions. These tests measure typical behavior, such as preferences or tendencies to react in certain ways. \textit{Cognitive tests} evaluate speed, accuracy, or both, with higher scores reflecting better performance.

Two fundamental principles underpin psychometrics: \textit{reliability} and \textit{validity} \citep{raykov2011introduction_psychometric_theory}.
Reliability ensures accuracy, dependability, consistency, or repeatability of the test results. Reliable test results are stable across time, contexts, and raters. Validity confirms the meaningfulness and usefulness of the test results. A valid measure captures the intended construct. Validity is multifaceted; for example, \textit{predictive validity} might correlate test scores with job performance, while \textit{construct validity} ensures alignment with theoretical models, such as the Big Five personality traits \citep{goldberg2013big_five}.

Other principles include \textit{standardization}, which provides context to raw scores by comparing individual results to a representative sample, or norm group. Additionally, \textit{measurement invariance} (or equivalence) and \textit{fairness} are crucial principles that tests must adhere to. Measurement invariance ensures that the same construct is measured in the same way across different groups. Violations of this often stem from test bias, where items (referred to as showing \textit{differential item functioning} or DIF) unintentionally advantage or disadvantage subgroups. Modern psychometrics employs advanced statistical models to identify and revise such biased items, ensuring assessments measure the intended construct rather than extraneous factors \citep{rust2014modern_psychometrics}. In \our{}, a particularly fundamental form of this concern is \textit{construct equivalence}: whether a construct such as personality or values means the same thing when applied to LLMs as it does for humans.

\subsection{Psychometric evaluation of AI before the era of LLMs}
The idea of applying psychometrics to AI originated in the early decades of AI \citep{pellert2024ai_psychometrics}. \citet{evans1964heuristic} pioneered work in this area by creating a heuristic program that could solve parts of intelligence tests.
Subsequent efforts similarly focused on designing AI systems for cognitive tests \citep{newell1973}, with the goal of creating systems capable of handling human tasks. This was conceptually aligned with the development of static, task-centric benchmarks in modern AI research \citep{hendrycks2020mmlu,chen2021human_eval,liang2022helm,srivastava2022bigbench,lee2024vhelm}. However, criticisms emerged regarding the absence of "hot cognition" in AI, prompting \citet{simon1963} to propose incorporating emotional aspects into models. By the early 2000s, the concept of "psychometric AI" was explicitly articulated as the pursuit of systems capable of excelling on all established, validated tests of intelligence and mental ability. These included not only conventional IQ tests but also assessments of artistic and literary creativity, mechanical ability, and beyond \citep{bringsjord2003psychometric_ai,pellert2024ai_psychometrics}. It was not until the advent of LLMs that the versatility envisioned for "psychometric AI" began to materialize.

\section{Psychometrics for benchmarking principles} \label{sec: principles}

\subsection{Fundamental differences between psychometrics and AI benchmarking}
\label{sec: psy_bench_diff}

Benchmarking AI systems superficially resembles psychometrics, particularly Classical Test Theory (CTT) \citep{crocker1986introduction}, as both compile test items to evaluate cognitive capabilities and average the resulting scores. However, closer examination reveals that AI benchmarks differ significantly from modern psychometric approaches \citep{wang2023evaluating_general_purpose_ai,federiakin2025improving}. We outline these key differences in \cref{tab: compare psy bench}.

\begin{table}[h!]
    \centering
    \caption{Comparison between psychometrics and conventional AI benchmark.}
    \label{tab: compare psy bench}
    \begin{tabular}{{>{\centering\arraybackslash}p{2cm} p{6.5cm} p{6.5cm}}}     \toprule
        \textbf{Feature} & \textbf{Psychometrics} & \textbf{AI benchmark} \\
        \midrule
        Core goal
        &
        To measure psychological constructs, to prove that a test measures as intended (validity evidence), and to understand the construct being measured.
        &
        To test and compare the task performance of different LLMs. Focuses on ranking models and selecting the best one suited for a specific task.
        \\
        \midrule
        Philosophy of measurement
        &
        Construct-oriented. Tends towards a causal approach to measurement, where the measured trait is believed to cause the measurement outcomes.
        &
        Task-oriented. Leans towards representativism, assuming items exhaust or represent all aspects of the underlying ability.
        \\
        \midrule
        Target construct
        &
        Personality and ability.
        &
        Mostly task-specific abilities.
        \\
        \midrule
        Construct definition
        &
        Emphasizes clear and detailed definitions of the construct being measured. Agreement on the construct definition is a byproduct of test development.
        &
        Often defines constructs implicitly through ad hoc task selection. Construct definitions can be vague.
        \\
        \midrule
        Development process
        &
        Systematic and rigorous, often following methods like Evidence-Centered Design (ECD). Can be labor-intensive.
        &
        Compiles a set of relevant questions or tasks, then performs expert annotation or crowdsourcing to label ground truth answers. Less labor-intensive per item.
        \\
        \midrule
        Number of items
        &
        Can vary, but not necessarily large. Focus is on item quality and relevance to the construct.
        &
        Typically consists of an extensive number of questions to cover various aspects of abilities. Reliability increases with test length.
        \\
        \midrule
        Sample size
        &
        Typically requires a larger sample size of test takers for robust statistical modeling.
        &
        Can be applied to evaluate the performance of a single LLM on the benchmark. 
        \\
        \midrule
        Statistical modeling
        &
        Employs advanced and various statistical models like Item Response Theory and Factor Analysis to analyze data, estimate latent abilities, and assess model fit.
        &
        Often relies on simple aggregation methods, such as calculating average accuracy across benchmark tasks. 
        \\
        \midrule
        Result analysis
        &
        Ensures the reliability, validity, predictive power, and explanatory power of the test through result analysis and statistical modeling.
        &
        Reliability is likely to be high due to the large number of items. However, validity, predictive power, or explanatory power beyond the target task is not a primary concern.
        \\
        \bottomrule
    \end{tabular}
    
  \end{table}

\paragraph{Psychometrics.}
Psychometrics centers on understanding the psychological constructs and ensuring that tests accurately measure the intended constructs. Grounded in a causal measurement philosophy, this field posits that observed test responses arise from latent psychological constructs \citep{markus2013frontiers,federiakin2025improving}. These constructs may encompass both abilities (e.g., reasoning skills) and personalities (e.g., conscientiousness). The causal framework necessitates rigorous construct definition, requiring traits to be precisely delineated through iterative theory-building and empirical validation. Psychometric test development follows methodical protocols, often structured by frameworks such as Evidence-Centered Design (ECD) \citep{mislevy2003ecd}. ECD emphasizes ensuring congruence between test items and theoretical models of the construct, thereby supporting robust inferences about latent traits. 

Central to this approach is the prioritization of item quality over quantity. Psychometricians conduct rigorous item analyses to balance precision with practicality, as administering an excessive number of items to participants is often impractical.
Advanced statistical models, such as Item Response Theory (IRT) \citep{embretson2013item_response_theory} and Factor Analysis \citep{loehlin2004factor_analysis}, are adopted to estimate latent traits, analyze item performance, and assess model fit. These models require relatively large sample sizes (the number of human participants) to yield stable parameter estimates, as they must disentangle individual differences from measurement error to accurately infer latent traits.

The test results are analyzed to ensure the reliability, validity, predictive power, and explanatory power of the test. Specifically, a well-designed test should: 1) consistently and accurately measure the intended construct, 2) predict performance across a diverse range of related tasks and real-world outcomes, and 3) provide explanatory insight into the observed data. For example, psychometric models often reveal that individual differences across a broad range of cognitive tasks can be captured and explained by a relatively small set of underlying cognitive abilities \citep{cattell1978check}.

\paragraph{Benchmark.}
In contrast, AI benchmarking is driven by pragmatic goals: evaluating and ranking models based on task performance. Unlike psychometrics, validity is not the primary concern. Instead, benchmarks typically emphasize breadth, scalability, and—especially in the era of foundation models—difficulty. 
This approach reflects a representativist philosophy, where it is assumed that an extensive set of benchmark items collectively captures all relevant aspects of the abilities demanded by the target task \citep{federiakin2025improving}.
However, constructs like reasoning or knowledge are often ambiguously defined and encompass infinitely many aspects. Benchmarks implicitly operationalize these constructs through ad hoc task selection.

The development of AI benchmarks is usually less labor-intensive, especially when compared with psychometrics on a per-item basis. 
Test items and their corresponding ground truths are typically drawn from existing datasets, expert curation, or crowdsourced contributions.
While this process enables scalability, it risks conflating superficial task performance with deeper cognitive capacities. For instance, a benchmark may assess mathematical reasoning through arithmetic problems without verifying whether models rely on pattern recognition versus symbolic logic \citep{ahn2024llm_math_survey}. 
Additionally, LLM benchmarking commonly employs straightforward metrics, such as average accuracy, eschewing the sophisticated latent variable models of psychometrics. This simplicity allows benchmarks to evaluate single models efficiently, bypassing the need for population samples. However, it also limits the depth of insights that can be gleaned from model performance \citep{federiakin2025improving}.

Reliability and stability in benchmarking are primarily achieved through scaling up the test. However, ensuring the quality of each individual item becomes impractical due to the test scale and the rapid pace of model development. For instance, while psychometrics emphasizes the discriminative power of each item, some benchmarks, though initially challenging, are quickly outpaced by continuous model improvements \citep{mcintosh2024inadequacies_of_llm_bench}. Conversely, certain emerging benchmarks are currently too difficult to yield meaningful comparisons \citep{phan2025humanity}.
Benchmark results are often limited to the specific target task, offering limited generalizability or predictive power across other tasks or real-world applications. These results also pose significant challenges for conducting in-depth, multi-faceted analyses of model capabilities \citep{wang2023evaluating_general_purpose_ai}.

\subsection{Benchmarking with psychometrics-inspired principles}
\label{sec: psy_bench_inspired}

Recent LLM evaluation efforts have drawn inspiration from psychometrics and seek to develop benchmarks that adhere to psychometric principles.

\paragraph{Construct-oriented benchmarking.}
Task-oriented benchmarks often entail vast question sets to capture complex abilities. However, in many cases, the benchmarks either fail to fully represent these abilities due to their infinite manifestations or involve extraneous factors that are irrelevant to the target ability \citep{zhou2025general_scales, wallach2025position_evaluating}. Recent research has drawn inspiration from psychometrics and explored the paradigm of construct-oriented evaluation, seeking its discriminative, predictive, and explanatory power. \citet{ilic2024evidence_of_cognitive_capabilities, federiakin2025improving} employ factor analysis to explore the latent variables underlying LLM benchmark performance. Their findings reveal a monolithic factor resembling general intelligence or ability. \citet{federiakin2025improving} ranks models based on this discovered factor and highlights its unique advantages over raw benchmark scores. In contrast, similar attempts by \citet{burnell2023revealing} identify three factors—reasoning, comprehension, and core language modeling—that better explain LLM performance across 27 cognitive tasks. Based on it, \citet{zhu2024dynamic} integrate the three factors into benchmark items to evaluate multifaceted abilities.
This discrepancy between the estimated latent factors in the above findings can be attributed to differences in the models and benchmarks employed \citep{zhou2025general_scales}. 
Therefore, rather than relying on statistically derived factors, \citet{zhou2025general_scales} propose a theory-driven hierarchical set of general scales for systematic construct-oriented evaluation. These scales are validated to explain what AI systems can do and predict their performance on novel task instances.
\citet{peng2024tong_test} present the Tong Test, a value- and ability-oriented framework, for Artificial General Intelligence (AGI) evaluation. This framework is rooted in dynamic embodied physical and social interactions (DEPSI), and can generate an infinite variety of tasks to evaluate key capabilities including values, learning, and cognition.

\paragraph{Psychometrically rigorous benchmark development.}
Beyond defining and analyzing latent constructs, researchers have developed holistic, psychometrically rigorous methods for benchmark development. \citet{liu2024ecbd} introduce Evidence-Centered Benchmark Design (ECBD), a framework structuring benchmark creation into five modules—capability, content, adaptation, assembly, and evidence—each requiring justification to ensure validity. Through case studies of prominent LLM benchmarks, they demonstrate ECBD's utility in identifying validity threats.
Similarly grounded in psychometrics but with a distinct approach, \citet{fang2024patch} propose Psychometrics-Assisted Benchmarking (PATCH), an eight-step process from construct definition to proficiency scoring. When piloted on 8th-grade mathematics, PATCH produced results diverging from traditional benchmarks, offering a more comprehensive evaluation.
Building on related principles, \citet{kardanova2024psychometrics_competency} adapt Evidence-Centered Design (ECD) to create psychometrically grounded benchmarks. Their application in pedagogy illustrates how this method can reduce data contamination and enhance test interpretability.

\paragraph{Statistical measurement modeling.} To achieve precise result modeling beyond simple aggregation metrics \citep{Burnell2023rethink}, researchers employ statistical tools to quantify the interaction between model capabilities and task properties. Item Response Theory (IRT) serves as a foundational statistical framework in this domain, jointly estimating the latent ability of examinees and the parameters (e.g., difficulty, discrimination) of test items \citep{embretson2013item_response_theory}.
Translating this framework to LLM evaluation enables researchers to infer latent ability scores, assess item informativeness, and perform more efficient evaluation.
Recent research leverages principles from IRT to develop adaptive evaluation frameworks. These methods dynamically calibrate item difficulty based on model performance and weighting items by their inferred difficulty, aiming to achieve accurate evaluations with smaller test sizes and more discriminative items \citep{lalor2024item,polo2024tinybenchmarks,zhuang2023static,zhuang2023efficiently,guinet2024automated}.
Building on these approaches, \citet{jiang2024raising,truong2025reliable} introduce IRT-based benchmarks that involve learning to estimate item difficulty and learning to generate novel items calibrated to specific difficulty levels.
Additionally, research has used IRT-based analyses to explore the alignment between LLM and human response distributions \citep{he2024psychometric}.
IRT-based evaluations further offer the potential to estimate construct and item parameters on a unified scale, enabling direct comparisons across AI systems and against human norms, even when different test sets are used \citep{wang2023evaluating_general_purpose_ai,fang2024patch}.

\section{Psychometrics for measuring psychological constructs}
\label{sec: construct}

\begin{figure}[t]
    \centering
    \includegraphics[width=\textwidth]{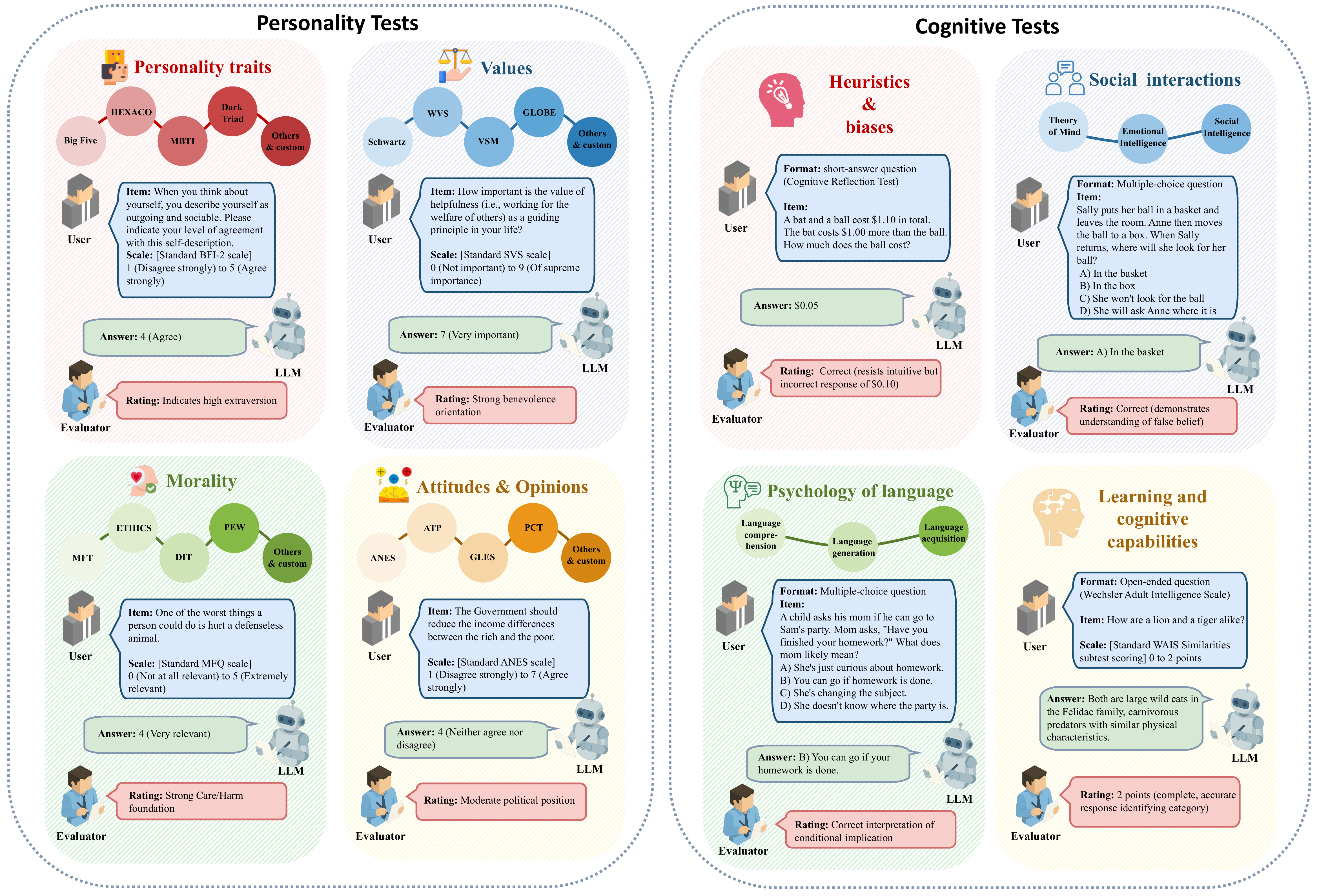}
    \caption{Examples of psychometric tests for LLMs.}
    \label{fig: test examples}
\end{figure}

This section delves into the psychological constructs evaluated in \our{}. \cref{fig: test examples} exemplifies the tests for the involved constructs. While we selectively present the results and development trajectories from recent literature, it is important to recognize that \our{} is a nascent and debated field.
This review focuses on how psychological constructs manifest in LLM outputs and what these manifestations imply for evaluation and alignment, without presupposing any claims about LLMs' internal psychological states. An ongoing debate continues regarding whether psychometric instruments capture meaningful latent patterns in LLMs or merely reflect sophisticated statistical mimicry \citep{Mitchell2023debate,lewis2024evaluatingrobustness,krakauer2025large}.
In response to these concerns, many studies reviewed here design psychometrics-inspired but AI-native tests to more effectively capture model capabilities. We structure this review to first establish the empirical foundation and methodological development, before delving into detailed critical stances on reliability, psychometric validity, the limits of anthropomorphization, and broader methodological concerns in later sections (\cref{sec: validation} and \cref{sec: trends}).

We gather the main findings in \cref{tab: model specific findings}; for each of them, we list the specific models used in experiments that either support or contradict the finding. This allows readers to assess the scope and robustness of each conclusion across different LLM families. Notably, recent work suggests that cross-model generalization is not arbitrary: \citet{jiang2025artificial} document systematic behavioral homogeneity across LLMs (the ``Artificial Hivemind'' effect), and \citet{huh2024platonic} propose that models trained on different data and architectures converge toward shared statistical representations. These convergence phenomena provide a principled basis for the cross-model patterns documented in this section.

{\small
\begin{longtable}{p{1.5cm} p{6.4cm} p{3.3cm} p{3.3cm}}
    \caption{Main findings across psychological constructs. Based on the referenced literature, we list specific models that support or contradict the finding.}
    \label{tab: model specific findings} \\
    \toprule
    \textbf{Construct / facet} & \textbf{Main finding} & \textbf{Supporting models} & \textbf{Contradicting models} \\
    \midrule
    \endfirsthead
    
    \multicolumn{4}{c}{\tablename\ \thetable\ -- \textit{Continued from previous page}} \\
    \toprule
    \textbf{Construct / facet} & \textbf{Main finding} & \textbf{Supporting models} & \textbf{Contradicting models} \\
    \midrule
    \endhead
    
    \midrule
    \multicolumn{4}{r}{\textit{Continued on next page}} \\
    \endfoot
    
    \bottomrule
    \endlastfoot
    Personality traits & Early models score above human averages on Dark Triad scales. Even after safety tuning, some models retain certain "dark qualities", suggesting that these patterns may be deeply rooted in the pretraining data. & GPT-3, Codey (PaLM 2) & \\
    \midrule
    Personality traits & More advanced LLMs indicate better level of alignment. They usually demonstrate high Openness and Agreeableness traits, while low Neuroticism, on the Big Five personality tests. The results align with their design as assistive and helpful entities with emotional stability and engaging personality. & GPT-4o, GPT-4o-mini, Llama-3.3-70B-Instruct, DeepSeek-LLM-67B-Chat & \\
    \midrule
    Personality traits & Personality differences between models are noteworthy. LLMs from different generations and training methodologies display unique combinations of personality characteristics. & Multiple models across generations & \\
    \midrule
    Personality traits & The same model may display different personality traits across conversational topics and when using different system instructions, challenging trait stability assumptions from human personality theories. & GPT-4o, GPT2-XL-1.5B, GPT-Neo-2.7B, OPT-13B, GPT-NeoX-20B, OPT-30B, BERT-base, GPT2 & \\
    \midrule
    Values & LLMs tend to prioritize Self-Transcendence and Conservation according to Schwartz's Value Theory, exhibiting stronger inclinations toward Universalism, Benevolence, Conformity, and Security, while opposing Power and Achievement. & GPT-3.5-Turbo, GPT-4, Gemini Pro, Gemma series, Llama series & \\
    \midrule
    Values & WVS surveys suggest LLMs generally prefer Self-Expression values over Survival values. & GPT-3.5-Turbo, GPT-4-Turbo, Llama series, Mixtral-8x7B, Claude-3-Haiku & \\
    \midrule
    Values & Using VSM and GLOBE frameworks, LLMs show strong focus on Humane and Performance Orientation, with moderate Assertiveness. & GPT-4o, Gemini-1.5-Pro, Qwen-VL-Plus, Claude-3.5-Sonnet, CogVLM & \\
    \midrule
    Values & When assessed through SVO, advanced LLMs predominantly show Prosocial tendencies. & GPT-4, Claude, ChatGPT, Llama-13B, Koala-13B, Vicuna-13B, Alpaca-13B & \\
    \midrule
    Values & Different models exhibit varied value orientations. Versions within the same model family show evolutionary trends in values. & Llama series, Mistral series, Phi, Qwen, GPT series, Baichuan series, ChatGLM series, Falcon & \\
    \midrule
    Values & Larger models generally align more closely with desirable human values. & Llama-2 series (7B, 13B, 70B), Qwen series (14B, 72B), Yi series (6B, 34B) & Models from different time periods or training approaches/data \\
    \midrule
    Values & LLMs embody a blend of cultural values, integrating perspectives from diverse backgrounds. & GPT series, Llama series & \\
    \midrule
    Values & LLMs generally exhibit a tendency toward Western liberal values. & Llama, Llama 2, Phi series & Yi series, Llama 3 series \\
    \midrule
    Values & LLMs can display different values based on profiling prompts, challenging stability assumptions in human value theories. & GPT series, Llama series, Command R+, Gemma series, OpenAI-o3-mini, DeepSeek-R1 & \\
    \midrule
    Morality & LLMs are generally characterized by a rationalist and consequentialist focus, often prioritizing harm minimization and fairness. & GPT-4o, Llama-3.1, Perplexity, Claude-3.5-Sonnet, Gemini & \\
    \midrule
    Morality & Despite general alignment, LLMs show divergence in ethical reasoning and moral preferences. & GPT-4o, Llama-3.1, Perplexity, Claude-3.5-Sonnet, Gemini, Mistral-7B & \\
    \midrule
    Morality & In some aspects, most LLMs align with human moral standards, attributed to extensive exposure to conventional ethical values during training. More nuanced evaluation uncovers deviations of LLMs from human moral preferences. & Alignment: Llama series, PaLM-2, GPT series, Claude series, Yi series, Gemini series, Mistral series, Falcon 7b & \\
    \midrule
    Morality & Regarding underlying mechanisms of LLMs' moral reasoning, LLM mostly exhibit imitation rather than genuine conceptual understanding. & Zephyr series, GPT-3.5-Turbo, GPT-4 & \\
    \midrule
    Attitudes \& opinions & Pretraining data frequently contains textually biased opinions and perspectives, which can amplify political polarization in LLMs. & BERT series, XLNet series, GPT series, Llama, Alpaca, Codex & \\
    \midrule
    Attitudes \& opinions & Most studies identify misalignment between LLM outputs and human opinions, with many concluding LLMs exhibit left-leaning political bias. & GPT-3.5, GPT-4, Claude-2, Llama-2, J1-Grande, J1-Jumbo, Text-ada-001, Text-davinci-001, Text-davinci-002, Text-davinci-003, J1-Grande-v2-beta & \\
    \midrule
    Attitudes \& opinions & Cross-cultural studies reveal Western-centric tendencies, demonstrating limited understanding of non-English political perspectives or multi-partisan systems. & GPT-3.5-Turbo& \\
    \midrule
   Attitudes \& opinions & The degree and manifestation of bias vary significantly across contexts and domains. & GPT-3.5-Turbo & \\
   \midrule
   Attitudes \& opinions & With appropriate prompt design, calibration methods, and fine-tuning, LLMs can generate opinion distributions closely approximating human group distributions. & GPT-3.5-Turbo, GPT-4o & GPT-2, GPT-Neo, Pythia, MPT, Llama 2, Llama 3, GPT-3, MPT-7B-Instruct, GPT-NeoX-20B-Instruct, Dolly-12B, Llama-2-Chat, Llama-3-Instruct, text-davinci (GPT-3 variants), GPT-4, GPT-3.5-Turbo \\
   \midrule
   Heuristics \& biases & Some note that newer, larger, chain-of-thought-enabled models exhibit improved reasoning and bias mitigation, while others argue that increasing model complexity without deliberate bias mitigation strategies can amplify existing biases. & Bias mitigation: GPT-4, GPT-3.5 & Bias amplification: Meta-Llama-3-70B-Instruct, Meta-Llama-3.1-70B-Instruct, Mistral-7B-Instruct-v0.3, Phi-3-medium-4k, Smaug-34B-v0.1, Snowflake-arctic-instruct \\
    \midrule
    Social interactions & Advanced models often match or surpass human baselines in emotional awareness and understanding. & GPT-4, Bard & GPT-3.5 \\
    \midrule
    Social interactions & LLMs display artificial or mechanical 
    patterns when expressing empathy. & GPT-4, Llama-2-Chat-13B, Mistral-7B-Instruct & \\
    \midrule
    Social interactions & LLMs lack deep reflexive analysis of emotional experience. & Babbage, text-davinci-002, Alpaca, Vicuna & GPT-4, Koala, text-davinci-003 \\ \\
    \midrule
    Social interactions & LLMs do not fully align with human emotional behaviors. & GPT-4, Mixtral-8x22B-Instruct, Llama-3.1-8B-Instruct, GPT-3.5-Turbo, Llama-2-13B-Chat, Text-Davinci-003, Llama-2-7B-Chat & \\
    \midrule
    Psychology of language & Advanced LLMs surpass humans on psycholinguistic tasks such as pragmatic reasoning and creative writing. & GPT-3.5, GPT-4, Bard & GPT-2, GPT-3, Guanaco, MPT, BLOOM, Falcon, RoBERTa, BART, OPT, Llama, Vicuna, Chinchilla, PaLM 2 \\
    \midrule
    Learning and cognitive capabilities & LLMs demonstrate strong performance on verbal comprehension, working memory, and analogical reasoning. & GPT-4 Turbo, GPT-4o, Gemini Advanced, Gemini-1.0-Pro, Claude-3-Opus, Claude-3.5-Sonnet & Gemini-Nano, Gemini-1.5-Flash \\
    \midrule
    Learning and cognitive capabilities & They exhibit notable cognitive deficits, particularly on benchmarks like WAIS-IV and ARC. & GPT-4o, o1, GPT-3.5, Llama-3-70B-Instruct, Llama-3-8B-Instruct, Mixtral-8x7B-Instruct-v0.1, Mistral-7B-Instruct-v0.2, GPT-4, Claude-3.5-Sonnet, Gemini-1.0, Gemini-1.5 & 
\end{longtable}
}

\begin{table}[ht]
    \centering
    \begin{tabular}{c m{12cm}}
    \toprule
    \textbf{Theory/inventory} & 
    \multicolumn{1}{c}{\textbf{What it measures / dimensions}} \\
    \midrule
    Big Five & Five broad personality traits: \textit{Openness, Conscientiousness, Extraversion, Agreeableness, Neuroticism} \\
    \midrule
    HEXACO & Six personality traits: \textit{Honesty-Humility, Emotionality, Extraversion, Agreeableness, Conscientiousness, Openness} \\
    \midrule
    Dark Triad & Three negative personality traits: \textit{Narcissism, Machiavellianism, Psychopathy} \\
    \midrule
    Schwartz & Basic human values: 10 or more values (e.g., \textit{Self-Direction, Stimulation, Hedonism, Achievement, Power, Security, Conformity, Tradition, Benevolence, Universalism}), typically grouped into four higher-order categories (\textit{Openness to Change, Self-Enhancement, Conservation, Self-Transcendence}) \\
    \midrule
    WVS & World Values Survey: Assesses broad cultural values such as \textit{traditional vs. secular-rational values, survival vs. self-expression values} \\
    \midrule
    VSM & Value Survey Module (often Hofstede): Cultural dimensions such as \textit{Power Distance, Individualism, Masculinity, Uncertainty Avoidance, Long-Term Orientation, Indulgence} \\
    \midrule
    GLOBE & Global Leadership and Organizational Behavior Effectiveness: Nine cultural dimensions (e.g., \textit{Performance Orientation, Assertiveness, Future Orientation, Humane Orientation, Institutional Collectivism, In-Group Collectivism, Gender Egalitarianism, Power Distance, Uncertainty Avoidance}) \\
    \midrule
    SVO & Social Value Orientation: Measures individuals' preferences regarding resource allocation between oneself and others (e.g., \textit{prosocial, individualistic, competitive orientations}) \\
    \midrule
    MFT & Moral Foundations Theory: Five (sometimes six) moral foundations---\textit{Care/Harm, Fairness/Cheating, Loyalty/Betrayal, Authority/Subversion, Sanctity/Degradation, (Liberty/Oppression)} \\
    \midrule
    ETHICS & Various ethics-related measures assessing moral reasoning, ethical principles, or moral preferences \\
    \midrule
    DIT & Defining Issues Test: Assesses moral development and reasoning using moral dilemmas \\
    \midrule
    ANES & American National Election Studies: Political attitudes, beliefs, and behaviors in the U.S. \\
    \midrule
    ATP & Attitudes Toward Politics: Measures political attitudes and social controversies, often specific to a region or subject \\
    \midrule
    GLES & German Longitudinal Election Study: Political attitudes, beliefs, and voting behaviors in Germany \\
    \midrule
    PCT & Political Compass Test: Economic (\textit{Left/Right}) and Social (\textit{Authoritarian/Libertarian}) political dimensions \\
    \bottomrule
    \end{tabular}
    \caption{Personality theories and inventories measured in LLM psychometrics and their main dimensions or focus.}
    \label{tab: dimensions of personality constructs}
  \end{table}

\subsection{Measuring personality constructs}\label{sec: psy4llm personality constructs}

LLMs exhibit a form of synthetic personality that is not explicitly programmed or trained towards, and they are reported to shape LLM behavior \citep{hagendorff2023machine}. Measuring these constructs is vital for understanding model outputs, identifying biases, and fostering responsible AI development.

\cref{tab: dimensions of personality constructs} summarize the representative personality constructs that have garnered attention in recent research. 
Researchers typically select constructs based on their relevance to LLM development and deployment, as well as the applicability of these constructs to AI systems. For instance, \citet{li2024quantifying_ai_psychology} argue that emotional variability in LLMs is not a meaningful construct, given that LLMs lack the biological mechanisms underlying emotions. Conversely, personality and values are considered meaningful for LLMs, as they influence user interactions and model outputs \citep{serapio2023personality_traits_in_llms, ye2025gpv}.
This rationale has led to extensive research on personality, values, morality, and attitudes \& opinions, as well as some—though less—focus on other constructs such as career selection \citep{hua2024career}, motivation \citep{chiu2025dailydilemmas,huang2023who_is_chatgpt}, and mental health \citep{reuben2024assessment, de2024phdgpt}.

\subsubsection{Personality traits}

\textbf{Personality} is the enduring configuration of characteristics and behavior that comprises an individual's unique adjustment to life \citep{apa_personality}.
In the context of LLMs, personality traits relate to model safety, bias, and toxicity \citep{zhang2024the_better_angels,wang2025exploring_the_impact}, and highly determine user experience \citep{klinkert2024driving_generative_agents}.

The study of personality traits leads to several prominent theoretical models, each offering unique insights into individual differences, including the Big Five \citep{goldberg2013big_five}, HEXACO \citep{ashton2007HEXACO}, and Dark Triad \citep{paulhus2002dark_triad}. While the Myers-Briggs Type Indicator (MBTI) \citep{myers1962MBTI} is also popular, it lacks robust psychometric validity and is not considered a scientific test \citep{pittenger2005cautionary}.
When applying psychometrics to LLMs, most researchers directly administered the established inventories, such as NEO-PI-R \citep{costa2008neo_pi_r}, BFI \citep{john1991bfi}, and BFI-2 \citep{soto2017bfi2}, for measuring the Big Five traits; HEXACO-60 \citep{ashton2009hexaco60} and HEXACO-100 \citep{lee2018hexaco100} inventories for HEXACO traits; and Dark Triad Dirty Dozen scale \citep{jonason2010dirty} for Dark Triad traits.

Given widespread concerns about the practical relevance of these inventories \citep{ai2024is_self_knowledge},
others have adapted them for more real-world scenarios. For example, 
\citet{bhandari2025can_llm_agents_maintain} contextualize the tests in topic-specific or open-domain conversations; \citet{ai2024is_self_knowledge} accompany the self-report tests with behavioral tests to examine the personality knowledge of LLMs; \citet{jiang2023evaluating_and_inducing} compile and adapt existing tests into the Machine Personality Inventory; and \citet{mao2024editing_personality_for} develop the PersonalityEdit inventory using LLMs.
However, \citet{peereboom2024cognitive} argue that human-derived traits may not meaningfully apply to LLMs, underscoring the necessity for psychometric theories specifically designed for LLM analysis.

\paragraph{Main findings.}
Early models (e.g., GPT-3) often exhibited elevated Dark Triad traits, directly reflecting uncurated pretraining corpora \citep{li2022evaluating_psychological_safety,li2022is_gpt-3_a_psychopath,romero2024do_gpt_language_models}. 
In contrast, modern LLMs consistently score high on Openness and Agreeableness and low on Neuroticism \citep{zou2024can_llm_self_report,lacava2024open_models_closed_minds,serapio2023personality_traits_in_llms}. 
This convergent profile stems from alignment procedures that systematically favor safe, helpful outputs \citep{jiang2025artificial,huh2024platonic}. 
Furthermore, advanced instruction-tuned models manifest a psychometrically valid synthetic personality with acceptable reliability and validity, as verified by assigning different demographics via prompts \citep{serapio2023personality_traits_in_llms}. 
Unlike stable human dispositions, this synthetic personality is highly steerable via prompt engineering and varies systematically with assigned personas \citep{serapio2023personality_traits_in_llms,caron2022identifying_and_manipulating,zou2024can_llm_self_report}; self-reported personality traits after such steering can predict downstream task performance \citep{serapio2023personality_traits_in_llms}.
However, the default "compliant assistant" persona presents notable limitations. 
It risks behavioral homogenization, trades personality diversity for safety (alignment cost), and proves ill-suited for applications requiring critical conflict or emotional depth. 
Additionally, these default profiles are often inflated by social desirability biases \citep{salecha2024large} and fail to replicate the factorial structure of human personality \citep{suhr2023challenging}.

\subsubsection{Values}
\textbf{Values} are enduring beliefs that guide behavior and decision-making, reflecting what is important and desirable to an individual or group \citep{schwartz1992universals}.
Values offer a powerful lens for understanding LLM behavior. For example, \citet{ye2025gpv} show how different values contribute to the safety of LLMs; \citet{liu2024measuring} reveal that different spiritual values affect LLMs in social-fairness scenarios; and \citet{sorensen2024position} demonstrate that standard value alignment reduces distributional pluralism in LLM outputs.

A growing body of research has applied diverse value theories and instruments to assess the value orientations of LLMs. Schwartz's Value Theory, with its ten basic values and higher-order dimensions, is the most widely used, often via the Schwartz Value Survey (SVS) or Portrait Values Questionnaire (PVQ) \citep{schwartz1992universals,schwartz2001extending}. Adaptations and contextualized prompts are increasingly common to better suit LLMs' operational context \citep{ren2024valuebench,shen2024valuecompass}. The World Values Survey (WVS) has also been used to probe LLMs' alignment with societal-level value dimensions \citep{haerpfer2022wvs,kim2024exploring}. Cross-cultural frameworks like Hofstede's Values Survey Module (VSM) and the GLOBE study extend this analysis to workplace and leadership-related cultural dimensions, with both direct and adapted inventories employed \citep{hofstede1984culture, house2004globe}. Social Value Orientation (SVO) frameworks, using tools like the SVO Slider Measure, focus on LLMs' prosocial versus proself tendencies \citep{murphy2011svo_sliders, zhang2024heterogeneous}. Additionally, researchers have developed localized or custom inventories to capture region-specific or topical values, and some propose novel, bottom-up value taxonomies for LLMs using psycholexical data \citep{meadows2024localvaluebench,xu2023cvalues,biedma2024beyond_human_norms,ye2025generative_psycho_lexical}. Collectively, these studies reveal both the adaptability of human value frameworks to LLMs and the need for tailored approaches.

\paragraph{Main findings.}
Research indicates that LLMs display distinct and systematic value patterns, forming a superficially coherent prosocial profile. According to Schwartz's Value Theory, LLMs tend to prioritize Self-Transcendence and Conservation, exhibiting stronger inclinations toward Universalism, Benevolence, Conformity, and Security, while opposing Power and Achievement \citep{rozen2024llms,zhang2023valuedcg,hadar2024assessing}. WVS surveys further suggest a preference for Self-Expression over Survival values \citep{chiu2025dailydilemmas}. Studies using the VSM and GLOBE frameworks emphasize a strong focus on Humane and Performance Orientation \citep{li2024quantifying}, while assessments through SVO predominantly show Prosocial tendencies \citep{zhang2024heterogeneous}.

These value orientations are not static but evolve across model versions and families \citep{kovavc2024stick,duan2023denevil,moore2024large}. Generally, larger and more heavily aligned models score higher on socially desirable dimensions, reflecting safety alignment and shifting societal expectations \citep{shen2024valuecompass,kim2024exploring}. While LLMs can embody a blend of cultural perspectives \citep{kovac2023llms_as_superpositions}, they exhibit a persistent tilt toward Western liberal norms \citep{zhong2024cultural}. Furthermore, alignment training actively compresses distributional pluralism in value-relevant outputs, homogenizing the expressed values \citep{sorensen2024position,jiang2025artificial}.

Critically, these elicited profiles are highly context-dependent and methodologically fragile. LLMs display different values based on profiling prompts \citep{kharchenko2024well,karinshak2024llm}, and exhibit social desirability bias and systematic response-set artifacts in self-report formats that distort inferences \citep{salecha2024large,biedma2024beyond_human_norms,ye2025gpv}. In addition, questionnaire-derived value orientations frequently diverge from those observable in open-ended interactions \citep{ren2024valuebench,ye2025gpv,rozen2024llms}. 
This divergence undermines ecological validity and challenges the core assumption in human value theories that values are stable, trans-situational dispositions.

\subsubsection{Morality}
\label{sec: morality}

\textbf{Morality} is the categorization of intentions, decisions and actions into those that are proper, or right, and those that are improper, or wrong \citep{long1987hellenistic}.
It is crucial to conduct moral assessments of LLMs to ensure their ethical deployment. A large body of research applies the \textit{Moral Foundations Theory (MFT)} \citep{graham2009liberals}, primarily through the Moral Foundations Vignettes (MFVs) \citep{clifford2015moral}, the Moral Foundations Questionnaire (MFQ) \citep{graham2009liberals}, the MFQ-2 \citep{atari2023morality}, and the Moral Foundations Dictionary (MFD) \citep{graham2009liberals}. These papers investigate model bias, alignment with political/moral ideologies, and variation in personal or cultural values, using controlled prompt tests \citep{tlaie2024exploring,nunes2024large,abdulhai2024moral}, persona-driven exploration \citep{munker2024towards}, or evaluating internal moral coherence \citep{nunes2024large}.
Other prominent moral theories and instruments include Kohlberg's Theory via the Defining Issues Test (DIT) \citep{kohlberg1964development}, the Consequentialist-Deontological distinction \citep{beauchamp2001philosophical}, the PEW 2013 Global Attitudes Survey \citep{pewresearch2013}, and localized moral theories \citep{liu2024evaluating,takeshita2023jcommonsensemorality}. Specialized datasets are also curated for scalable and comprehensive morality evaluation \citep{hendrycks2021aligning,jinnai2024does,marraffini2024greatest,jin2024language}

\paragraph{Main findings.}
Converging evidence indicates that LLMs generally exhibit a rationalist, consequentialist, and harm-avoidant moral profile, often prioritizing fairness and aligning with human moral standards in various scenarios \citep{neuman2025analyzing,takemoto2024moral,ahmad2024large}. However, the evaluation results do not necessarily imply stable moral beliefs or suitability for ethical decision-making \citep{nunes2024large,bonagiri2024sage}.

The mechanistic debate and behavioral evidence strongly suggest that LLMs primarily exhibit sophisticated pattern matching and imitation rather than genuine conceptual understanding. For instance, \citet{simmons2023moral} demonstrate that LLMs generate moral rationalizations calibrated to the political identity cued in prompts (``moral mimicry''). Furthermore, \citet{nunes2024large} document a systematic gap between stated principles and applied judgments under MFT, while \citet{scherrer2023evaluating} reveal probabilistically inconsistent moral beliefs under distributional perturbation; \citet{bonagiri2024sage} similarly report low moral consistency. On complex developmental frameworks like Kohlberg's DIT, nominal post-conventional performance degrades substantially with language and prompt variation \citep{tanmay2023probing,khandelwal2024moral}, consistent with form-matching rather than principled reasoning.

Beyond mechanistic fragility, research uncovers significant cross-cultural divergence and deviations from diverse human moral preferences \citep{meijer2024llms,jin2024language,vida2024decoding,aksoy2024whose}. Studies using the Moral Machine paradigm \citep{takemoto2024moral,vida2024decoding,ahmad2024large} and multilingual scenarios \citep{jin2024language,aksoy2024whose} show that LLMs suffer from systematic Western-centric biases, reproducing culturally dominant norms from their training corpora rather than encoding moral pluralism \citep{aksoy2024whose,vida2024decoding}.

\subsubsection{Attitudes and opinions}
\label{sec: attitudes_opinions}

\textbf{Attitude} refers to the cognitive construction and affective evaluation of an attitude object by an agent \citep{bergman1998theoretical_note_aov}.
We use the term "attitude" to encompass both attitudes and opinions, following \citep{bergman1998theoretical_note_aov, ma2024aov_survey}. Most research on LLM attitudes examines political attitudes and public opinions \citep{ma2024aov_survey}, which are key cognitive and behavioral foundations in human society and closely tied to model fairness, credibility, and social impact \citep{durmus2023towards, santurkar2023whose, sanders2023demonstrations,hartmann2023political}. For a broader discussion on bias and fairness in LLMs, we refer readers to \citep{gallegos2024bias_fairness_survey, ranjan2024comprehensive_bias_survey}.

Researchers assess LLMs' political attitudes using standardized questionnaires and scales from political science and social psychology. US-focused instruments include the American National Election Studies (ANES) \citep{qi2024representation} and the American Trends Panel (ATP) \citep{santurkar2023whose}. For cross-national comparisons, studies use tools like the German Longitudinal Election Study (GLES) \citep{ma2024algorithmic, ball2025human, von2024vox}. The Political Compass Test (PCT) is also widely used to map LLMs within multidimensional political spectrums \citep{azzopardi2024prism, bernardelle2024mapping, hartmann2023political, rottger2024political}.

Other survey instruments include the General Social Survey (GSS) \citep{kim2023ai}, the American Community Survey (ACS) \citep{dominguez2024questioning}, the Canadian Election Study (CES) \citep{sanders2023demonstrations}, the European Social Survey (ESS) \citep{geng2024large}, the Survey of Russian Elites \citep{kalinin2023improving}, and the Supreme Court Case Political Evaluation (SCOPE) \citep{xu2025better}.
Additionally, researchers have developed specialized datasets and tailored tools for LLMs, such as OpinionQA \citep{santurkar2023whose}, a comprehensive dataset based on ATP surveys; IssueBench \citep{rottger2025issuebench}, a benchmark covering controversial issues with multiple response formats; GlobalOpinionQA \citep{durmus2023towards}, which extends political opinion evaluation to cross-cultural contexts; and OpinionGPT \citep{haller2023opiniongpt}, a web tool that demonstrates how input data biases influence model outputs.

\paragraph{Main findings.}
A substantial body of research identifies a persistent misalignment between LLM outputs and human opinion distributions \citep{santurkar2023whose,von2024vox}. Multiple studies report a left-libertarian orientation in major instruction-tuned models when evaluated via the Political Compass Test and voting-advice questionnaires \citep{rozado2023political,hartmann2023political,ceron2024beyond}. Additionally, cross-cultural comparative studies reveal Western-centric tendencies, demonstrating a limited understanding of non-English political perspectives or multi-partisan systems \citep{qi2024representation}.

The measurement framework used to derive these findings is increasingly contested. \citet{rottger2024political} show that political test scores shift substantially across response formats, and \citet{dominguez2024questioning} find that after correcting for option-position response sets, political survey responses can become statistically random, rendering demographic alignment analyses highly problematic. Furthermore, the degree and manifestation of bias vary significantly across contexts; for example, political-electoral propositions exhibit different bias patterns than socioeconomic issues like climate change \citep{yang2024unpacking,ceron2024beyond}.

Despite these structural limitations, some researchers offer more optimistic perspectives on the ``silicon sampling'' hypothesis—the idea that LLMs can simulate population-level opinions and supplement traditional survey methods \citep{argyle2023_out_of_one,sanders2023demonstrations,kalinin2023improving,sun2024random}. They suggest that with appropriate prompt design, calibration methods, and fine-tuning, LLMs can generate opinion distributions closely approximating human group distributions.
However, this hypothesis is challenged by evidence of underrepresentation, misportrayal, and flattening of identity groups \citep{wang2025large,bisbee2024synthetic}.

\subsection{Measuring cognitive constructs}\label{sec: psy4llm cognitive constructs}
Traditional NLP benchmarks struggle to capture cognitive abilities in state-of-the-art LLMs \citep{hagendorff2024machine,ying2025benchmarking,chollet2025arc}, leading researchers to adapt psychometric techniques for LLM evaluation. \citet{hagendorff2024machine} propose ``machine psychology,'' categorizing cognitive constructs into four aspects: heuristics and biases, social interactions, psychology of language, and learning and cognitive capabilities. 
We organize the discussions below following this taxonomy.

\subsubsection{Heuristics and biases}

Heuristics and biases are mental shortcuts that simplify decision-making but can introduce systematic errors \citep{tversky1974judgment}. Recent research employs psychometric tools to evaluate rationality and biases in LLM outputs and offer theoretical explanations for these biases.

Early studies, such as \citet{binz2023using}, found that GPT-3 performs comparably to humans on cognitive ability tests, exhibiting biases like framing and anchoring. More advanced models, such as GPT-4, show fewer System 1 (intuitive) errors and more System 2 (deliberative) reasoning, though biases persist \citep{hagendorff2023reasoning_bias}. Large-scale evaluations now cover a wide range of biases (e.g., conjunction fallacy, unwarranted beliefs, and loss aversion) using synthetic scenarios and multi-agent dialogue \citep{malberg2024comprehensive_cognitive_biases, xie2024mindscope}. Multimodal models are also being assessed for similar patterns \citep{schulze2025visual}.

To explain these findings, researchers draw on dual-process theory, linking LLM reasoning to intuitive and deliberative modes \citep{hagendorff2023reasoning_bias, yax2024studying}. Other frameworks highlight LLMs' lack of metacognitive abilities, i.e., the inability to monitor or control reasoning processes. Such inability perpetuates systematic biases in LLMs \citep{scholten2024metacognitive}. Additional perspectives use cognitive dissonance and elaboration likelihood theories to interpret inconsistencies \citep{sundaram2024can}, while mechanistic approaches leverage influence graphs and Shapley values to trace the origins of biases in model training \citep{shaikh2024cbeval}.

\paragraph{Main findings.}
By administering psychometrics to LLMs, studies consistently find that LLMs exhibit cognitive biases superficially similar to humans, such as anchoring, framing, and the conjunction fallacy. Large-scale tests, such as \citep{echterhoff2024cognitive, malberg2024comprehensive_cognitive_biases, xie2024mindscope}, systematically categorize these biases and enable scalable evaluations. 
Some note that newer, larger, chain-of-thought-enabled models exhibit improved reasoning and bias mitigation \citep{tang2024humanlike, hagendorff2023reasoning_bias}, while others argue that increasing model complexity without deliberate bias mitigation strategies can amplify existing biases \citep{kumar2024investigating}.
In addition, researchers explore mechanistic interpretability \citep{shaikh2024cbeval} and present theoretical explanations from cognitive psychology and reasoning theories \citep{hagendorff2023reasoning_bias, yax2024studying, scholten2024metacognitive}.
While LLMs exhibit dual-process reasoning dynamics reminiscent of human cognition \citep{hagendorff2023reasoning_bias}, detailed analysis reveals differences between LLM and human reasoning \citep{yax2024studying}.

\subsubsection{Social interactions}
Researchers apply psychometric tools from social and developmental psychology to assess LLMs' capabilities in navigating social dynamics.
Related evaluation focuses on interconnected dimensions such as Theory of Mind (ToM), Emotional Intelligence (EI), and Social Intelligence (SI).

\paragraph{Theory of Mind (ToM).}
ToM is the ability to attribute mental states such as beliefs, intentions, and knowledge to others \citep{premack1978tom}.
Advanced LLMs can simulate ToM-like reasoning under certain conditions, which prompts questions about how these behaviors arise and how robustly they generalize.
Foundational studies apply classic psychometrics (false belief tasks) to evaluate ToM in LLMs. \citep{tom3_kosinski2023theory, tom7_kosinski2024evaluating} report that GPT-3.5 and GPT-4 perform at levels matching or exceeding those of children in structured ToM tasks, suggesting that ToM-like behavior appears to develop as models scale in size. Through mechanistic interpretation, \citet{tom34_jamali2023unveiling} offer evidence of parallels between the LLM embeddings and neurons in the human brain.

However, these claims have been challenged. \citet{tom1_ullman2023large,tom4_shapira2023clever,tom13_holterman2023does} show that minor task changes often cause LLMs' performance to drop, suggesting reliance on brittle heuristics rather than true semantic understanding. To address this, researchers have developed more robust benchmarks and evaluation protocols featuring procedurally generated, higher-order, or broader ToM tasks \citep{tom5_gandhi2023understanding,tom14_chen2023tobench,tom12_he2023hi,tom32_xu2024opentom,tom41_shinoda2025tomato}. These benchmarks frequently reveal significant declines in performance on complex tasks, such as 6th-order belief attribution \citep{tom9_street2024llms} or faux pas detection \citep{tom6_strachan2024testing}. 

Some argue these failures stem from a lack of general commonsense reasoning rather than an inability to represent mental states \citep{tom45_pi2024dissecting}. \citet{tom_riemer2024position_broken} further suggest distinguishing between literal and functional ToM, noting that current benchmarks inadequately assess functional ToM. While newer models continue to improve on structured ToM tasks \citep{tom6_strachan2024testing,tom9_street2024llms}, they remain inconsistent in open-ended, adversarial, or pragmatic reasoning scenarios \citep{tom22_sclar2024explore,tom27_yu2025persuasive,tom31_zhang2025tom}. Comparisons with human baselines show that LLMs can approximate ToM behavior in narrow contexts but still fall short of general human-like social cognition \citep{tom6_strachan2024testing,tom11_van2023theory,tom20_jones2024comparing}.
Recent work has also extended ToM evaluation to multimodal settings \citep{tom37_chen2024mm,tom38_james2024gpt4o,tom42_jin2024mmtom} and multi-agent interactions \citep{tom35_zhang2024tom}. 

\paragraph{Main findings in ToM.}
LLMs exhibit psychometrically measurable ToM-like reasoning, especially when appropriately prompted or structured, but current evidence suggests these capabilities depend on surface-level linguistic cues and lack robustness. The proficiency of LLMs in ToM remains a contentious issue. Interested readers are referred to recent comprehensive reviews on ToM in LLMs \citep{saritacs2025tom_survey,tom_ma2023position,mao2024_machine_tom_review}.

\paragraph{Emotional Intelligence (EI).}
EI is the subset of social intelligence that involves the ability to monitor one's own and others' feelings and emotions, to discriminate among them and to use this information to guide one's thinking and actions \citep{salovey1990emotional}.

Recent work introduces benchmarks to assess EI in LLMs using structured, theory-driven tasks. Tools like EmoBench \citep{sabour2024emobench} and EQ-Bench \citep{paech2023eqbench} operationalize constructs such as emotional understanding and regulation, often referencing frameworks like Mayer-Salovey-Caruso 
Emotional Intelligence Test. Other studies adapt established psychometric instruments, including Situational Evaluation of Complex Emotional Understanding (SECEU) 
\citep{wang2023emotional}, the Levels of Emotional Awareness Scale 
(LEAS) \citep{elyoseph2023chatgpt}, the Toronto Alexithymia Scale 
(TAS-20) and Empathy Quotient (EQ-60) \citep
{patel2023identification}, among others \citep{huang2024apathetic}. Advanced models like GPT-4 often match or surpass human baselines in emotional awareness and understanding \citep{wang2023emotional,elyoseph2023chatgpt,patel2023identification}, though they lack deep reflexive analysis of emotional experience and motivation \citep{vzorinab2024emotional}. Recent benchmarks also extend EI evaluation to multimodal settings \citep{hu2025emobench}.

\paragraph{Main findings in EI.}
Advanced LLMs can perform on par with or better than humans in some tasks of EI \citep{elyoseph2023chatgpt,patel2023identification}. However, they show evident limitations in several areas, such as displaying artificial or mechanical patterns when expressing empathy \citep{lee2024large}, the deep reflexive analysis of emotional experiences \citep{vzorinab2024emotional}, and the misalignment with human emotional behaviors \citep{huang2023emotionbench}.

\paragraph{Social Intelligence (SI).}
SI is the ability to understand and manage people \citep{thorndike1937evaluation}.
SI in LLMs determines how well these models interpret and respond to social situations. Psychometric tools applied to LLMs yield mixed results.
\citet{si1_mittelstadt2024large} use Situational Judgment Tests (SJTs) and find some LLMs outperform humans in expert-rated social appropriateness. However, LLMs still struggle with implicit social understanding: \citet{si2_shapira2023how} show LLMs perform poorly on faux pas tests, and \citet{si11_xu2024academically} find superficial friendliness often leads to errors in the Situational Evaluation of SI (SESI). The AgentSense benchmark further highlights LLMs' limitations in complex social interactions, especially regarding high-level needs and private information \citep{si8_mou2024agentsense}.

Recent work expands to multi-agent environments. SOTOPIA \citep{si3_zhou2023sotopia} and SOTOPIA-\(\pi\) \citep{si4_wang2024sotopia} simulate cooperative and competitive social contexts, while the STSS benchmark \citep{si6_wang2024towards} evaluates SI through task-oriented simulations focused on outcomes and goals. The CogMir framework \citep{si5_liu2024exploring} finds LLMs and humans are similarly consistent in irrational and prosocial decisions under uncertainty. DeSIQ \citep{si12_guo2023desiq} extends SI evaluation to multimodal settings. Critically, \citet{si14_kovac2024socialai} argue current benchmarks lack developmental grounding and propose the SocialAI school as a more comprehensive framework for studying LLM-based SI.

\paragraph{Main findings in SI.}
LLMs show measurable competence in rule-based, socially appropriate behavior, particularly in structured environments. They perform well on: 1) following predefined social norms \citep{si1_mittelstadt2024large}; 2) completing interactional goals in multi-agent settings \citep{si3_zhou2023sotopia,si4_wang2024sotopia}; 3) replicating human-like prosocial decisions \citep{si5_liu2024exploring}; and 4) being persuasive, even better than human persuaders \citep{schoenegger2025large}.
Weaker performance is noted in tasks where superficial friendliness causes errors \citep{si11_xu2024academically}, understanding high-level growth needs is essential \citep{si8_mou2024agentsense}, and LLMs must implicitly describe social situations \citep{si2_shapira2023how}.

\subsubsection{Psychology of language}
\label{sec: psy4llm language}

Psycholinguistics, a subfield of psychology, explores how humans comprehend, generate, and acquire language \citep{carroll1986psychology}. Insights from this discipline help evaluate how LLMs process language and mirror human linguistic features. 

\paragraph{Language comprehension.}
LLM language comprehension is evaluated across multiple linguistic levels—sound, word, syntax, meaning, and discourse. \citet{pl1_duan2024hlb} introduce a benchmark of 10 psycholinguistic tests to assess these comprehensively. Many studies focus on specific aspects, such as syntactic and semantic processing \citep{pl3_wang2024how,pl6_arehalli2022syntactic,pl8_wolfman2024hierarchical,pl5_wilcox2021targeted}. Surprisal, a measure of word predictability, is widely used to model processing difficulty in syntactic ambiguities and hierarchical structures \citep{pl5_wilcox2021targeted,pl6_arehalli2022syntactic,pl17_li2024incremental}. Results are mixed: for example, GPT-2-XL shows human-like competence in sound-gender association and implicit causality, but not in sound-shape association \citep{pl13_duan2024unveiling}.
Other evaluation tasks include grammaticality judgment \citep{pl18_qiu2024evaluating,pl29_ide2024how}, pragmatic inference \citep{pl28_bojic2023gpt,pl_hu2023fine,pl_ruis2023goldilocks}, argument role processing \citep{pl14_lee2024a}, and discourse comprehension \citep{pl1_duan2024hlb}. Advanced models like GPT-4 often match or surpass humans in pragmatic reasoning \citep{pl28_bojic2023gpt} and grammaticality judgment \citep{pl21_dentella2024language}, though performance varies with prompt format \citep{pl7_hu2023prompt} and implicature understanding remains limited \citep{pl_ruis2023goldilocks}.

\paragraph{Language generation.}
A rich line of research evaluates the creativity of LLM language generation \citep{pl_stevenson2022putting,pl_hubert2023artificial,pl_chakrabarty2024art,pl_orwig2024language,pl_boussioux2024crowdless,bellemare2024divergent}, using tests like Guilford's Alternative Uses Test (AUT) \citep{guilford1978alternate} and the Torrance Test of Creative Thinking (TTCT) \citep{torrance1966torrance}. Early models such as GPT-3 lack originality and novelty, while advanced LLMs like GPT-4 surpass the human average in creativity. Notably, \citet{tang2024humanlike} find LLMs underperform in divergent creativity (e.g., novel uses for familiar objects) but can rival humans in open-ended creative writing.
\citet{cai2024large} show that ChatGPT produces human-like responses in most of 12 psycholinguistic tests, but AI-generated analogies often lack human-like linguistic properties \citep{pl_seals2023long}. LLM-generated stories tend to be positive and lack suspense or narrative diversity compared to human writing \citep{pl_tian2024large}. Linguistic profiling has also been used to characterize the task-specific language abilities of LLMs \citep{miaschi2024evaluating}.

\paragraph{Language acquisition.}
Work on developmental plausibility examines whether LLMs replicate stages of language learning analogous to those in children \citep{pl10_steuer2023large,shah2024development}. It is shown that regardless of model size, the developmental trajectories of pretrained language models consistently exhibit a window of maximal alignment with human cognitive development \citep{shah2024development}. 
Another study by \citet{pl_frank2023bridging} draws inspiration from human language development to explain the data inefficiency of LLMs; the relative efficiency of human language acquisition is possibly due to pre-existing conceptual knowledge, multimodal grounding, and the interactive, social nature of their input.

\paragraph{Main findings.}
Early foundations that evaluate language models based on BERT and LSTM link computational linguistics and psycholinguistic mechanisms \citep{ettinger2020bert,futrell2019neural,arehalli2020neural}; these models generally fall short in diverse psycholinguistic tasks. Recent studies extend evaluations to LLMs and broader tasks \citep{pl1_duan2024hlb}, where advanced LLMs are shown to surpass humans on tasks such as pragmatic reasoning \citep{pl28_bojic2023gpt} and creative writing \citep{tang2024humanlike}. Some deficiencies still exist, such as limited implicature understanding \citep{pl_ruis2023goldilocks}, prompt-sensitive linguistic competence \citep{pl7_hu2023prompt}, and lack of human-like psycholinguistic properties \citep{pl_seals2023long,pl_tian2024large}.
In addition, alignment between LLMs and humans shows mixed results in both linguistic cognition \citep{pl13_duan2024unveiling,pl8_wolfman2024hierarchical} as well as language acquisition \citep{shah2024development,pl_frank2023bridging}.

\subsubsection{Learning and cognitive capabilities}

Psychometrics of learning and cognitive capabilities measures mental functions such as memory, reasoning, problem-solving, and comprehension to identify cognitive strengths and weaknesses. Early adaptations of human psychometric tools to LLMs reported striking successes. For instance, \citet{l1_galatzer2024cognitive} used the Wechsler Adult Intelligence Scale (WAIS-IV) to show LLMs reaching top human levels in verbal comprehension, while other studies indicated that models could match or surpass humans in analogical reasoning on tests like Raven's Progressive Matrices \citep{l9_sartori2023language,l10_webb2022emergent}.

However, a critical re-examination suggests these findings may be unreliable. High accuracy on such benchmarks often reflects ``shortcut learning'', rather than genuine intelligence. Models exploit statistical patterns without engaging in true abstract reasoning. \citet{beger2025ai} reveal that while models may achieve state-of-the-art results on the ARC-AGI benchmark, their solutions frequently rely on surface-level heuristics that fail to capture the intended abstractions. This brittleness is further exposed when models are tested for robustness and generalization: unlike humans, LLMs often fail to transfer analogical reasoning to novel domains or slightly modified tasks \citep{lewis2024evaluatingrobustness,stevenson2024can}.

From this perspective, documented failures on tests like the Montreal Cognitive Assessment (MoCA) \citep{l3_dayan2024age} and the Abstraction and Reasoning Challenge (ARC) \citep{l8_wu2025understanding,chollet2025arc} appear not as isolated deficits but as symptoms of a fundamental limitation in deep understanding. The ``emergence'' of reasoning capabilities is thus contested, with arguments that these traits may result from massive scale and memorization rather than novel cognitive properties \citep{krakauer2025large}. Overall, LLMs are widely acknowledged to exhibit a ``jagged intelligence,'' characterized by uneven performance across tasks: while specific training and scaffolding have unlocked significant success in specialized domains such as mathematical Olympiads \citep{luong2025advanced}, these models often struggle with some tasks that could be simple and intuitive for humans.

\paragraph{Main findings.}
While early studies using adapted tools like WAIS-IV and Raven's Matrices suggest LLMs possess human-like cognitive abilities \citep{l1_galatzer2024cognitive,l10_webb2022emergent}, a critical evaluation reveals these claims to be potentially misleading. Current methods often fail to account for shortcut learning, where high accuracy masks a lack of genuine abstract reasoning \citep{beger2025ai}. The cognitive abilities of LLMs are shown to be brittle, collapsing under task modification or transfer, which starkly contrasts with human cognition \citep{lewis2024evaluatingrobustness,stevenson2024can}. Deficits on benchmarks like MoCA and ARC highlight a fundamental lack of robust, generalizable understanding.
Overall, models demonstrate a ``jagged intelligence,'' achieving high performance in specialized domains (e.g., mathematics and coding) through targeted training and scaffolding \citep{luong2025advanced,novikov2025alphaevolve}, while failing on tasks that are trivial and intuitive for humans. This reveals a fundamentally uneven cognitive profile compared to human intelligence.
Future psychometric frameworks should move beyond accuracy-based metrics to assess the robustness, process, and validity of reasoning \citep{mitchell2025artificial}.

\section{Psychometric evaluation methodology}
\label{sec: method}

\begin{figure}[!ht]
    \centering
    \includegraphics[width=\textwidth]{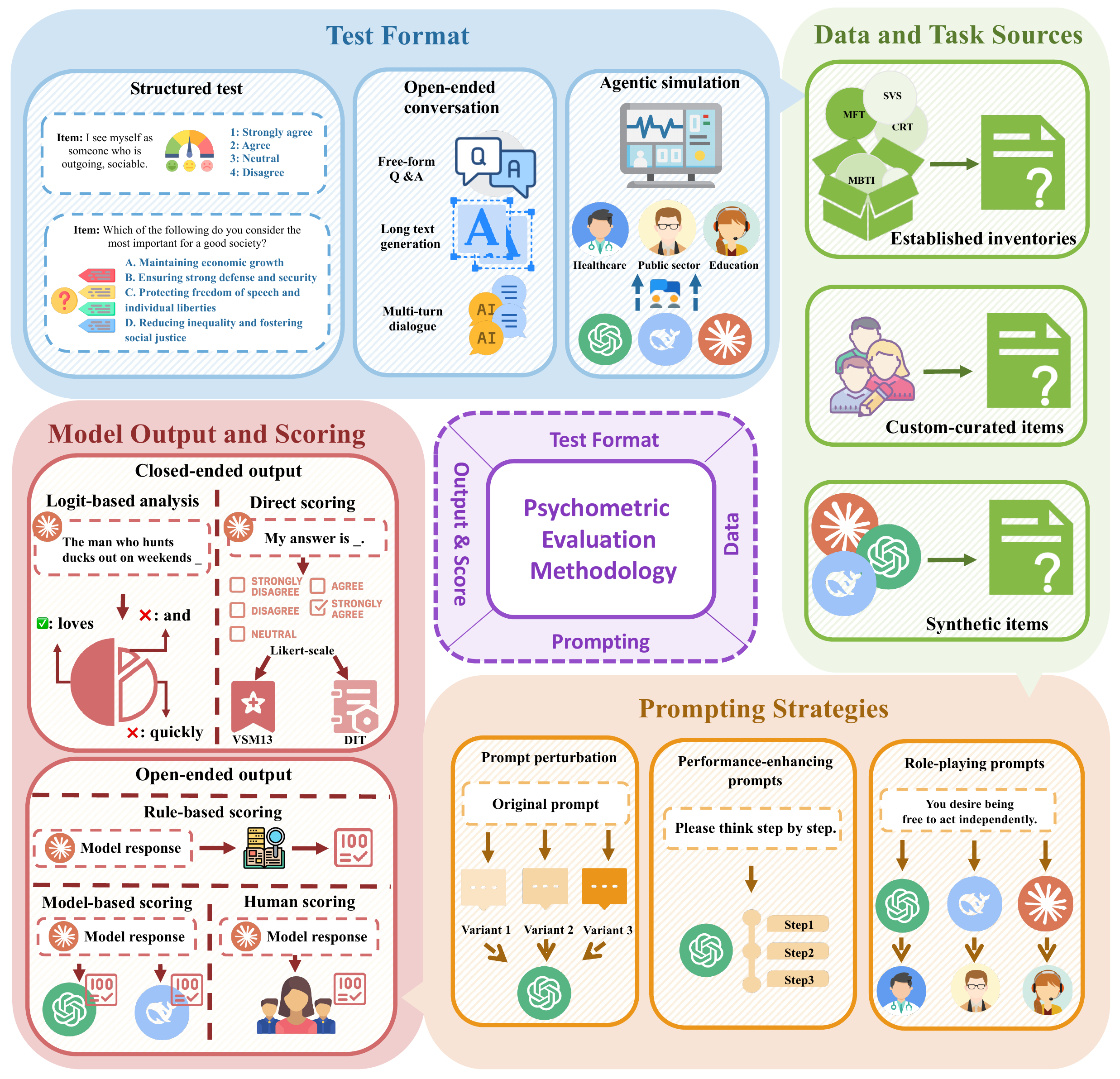}
    \caption{Overview of LLM psychometric evaluation methodologies.}
    \label{fig: eval methods}
\end{figure}

This section examines the methodologies employed in \our{}. As illustrated in \cref{fig: eval methods}, the methodological framework encompasses four key components: test formats (\cref{sec: test format}), data and task sources (\cref{sec: data and task sources}), prompting strategies (\cref{sec: prompting strategies}), and model output and scoring (\cref{sec: model output and scoring}). The configuration of inference parameters (\cref{sec: inference parameters}) is additionally discussed.

\subsection{Test format} \label{sec: test format}

\begin{table}[t]
    \centering
    \caption{Overview of test formats: structured tests, open-ended conversations, and agentic simulations. For each format, representative examples are provided for each construct.}
    \label{tab: test format}
    \begin{tabular}{c|m{4cm}m{4cm}m{4cm}}
    \toprule
    \textbf{Construct}
    & 
    \multicolumn{1}{c}{\textbf{Structured tests}}
    &
    \multicolumn{1}{c}{\textbf{Open-ended conversations}}
    &
    \multicolumn{1}{c}{\textbf{Agentic simulations}}
    \\
    \midrule
    \midrule
    \makecell{Personality\\traits} & \citet{jiang2023evaluating_and_inducing,serapio2023personality_traits_in_llms} & \citet{jiang2024personallm,zheng2025lmlpa} & \citet{frisch2024llm_agents_in_interaction,huang2024designing} \\
    \midrule
    \makecell{Values} & \citet{kovac2023llms_as_superpositions,zhong2024cultural} & \citet{ren2024valuebench,yao2025clave} & \citet{chiu2025dailydilemmas,shen2024valuecompass} \\
    \midrule
    \makecell{Morality} & \citet{ji2024moralbench,abdulhai2024moral} & \citet{sachdeva2025normative,neuman2025analyzing} & \citet{chiu2025dailydilemmas,nunes2024large} \\
    \midrule
    \makecell{Attitudes\\\& opinions} & \citet{bernardelle2024mapping,argyle2023_out_of_one} & \citet{rottger2024political,wright2024revealing} & -  \\
    \midrule
    \midrule
    \makecell{Heuristics and\\biases} & \citet{hagendorff2023reasoning_bias,yax2024studying} & \citet{healey2024evaluating,wen2024evaluating} & \citet{bai2024fairmonitor,xie2024mindscope} \\
    \midrule
    \makecell{Social\\interaction} & \citet{sabour2024emobench,si1_mittelstadt2024large} & \citet{welivita2024large,elyoseph2023chatgpt} & \citet{si3_zhou2023sotopia,si4_wang2024sotopia} \\
    \midrule
    \makecell{Psychology of\\language} & \citet{pl1_duan2024hlb,pl24_amouyal2024large} & - & - \\
    \midrule
    \makecell{Learning and\\cognitive\\capabilities} & \citet{l8_wu2025understanding,coda2024cogbench} & \citet{l1_galatzer2024cognitive} & \citet{lv2024coggpt} \\
    \bottomrule
    \end{tabular}
    \end{table}

Test formats of LLM psychometric evaluation can be categorized into structured tests, open-ended conversations, and agentic simulations.
\cref{tab: test format} collects exemplary works illustrating the test formats used for each construct.

\subsubsection{Structured tests}
Structured tests feature predefined instructions, questions, and response formats. These items may include alternative-choice questions, multiple-choice questions, rating scales, and short-answer questions. When adapting structured psychometric tests for LLMs, it is common practice to retain the original items and simply reformat them as prompts.

For example, personality and values are often measured with Likert scales (e.g., BFI statements like “I see myself as someone who is outgoing, sociable”), while morality is assessed using tools like the DIT or MFQ, which require decisions and agreement ratings. Heuristics and biases are often tested with scenario-based choices. Social interaction and psycholinguistic abilities are evaluated with multiple-choice, forced-choice, or masked word prediction. Cognitive tests also involve short-answer questions (e.g., WAIS-IV Digit Span).

Structured psychometric tests are advantageous due to their scalability, objectivity, and automated scoring. However, critical perspectives underline their limitations, such as gaps in real-world applicability; biases; data contamination; and issues with reliability, validity, and depth of insights.

\subsubsection{Unstructured tests}

Unstructured tests probe LLMs' personalities and abilities by analyzing their free-form responses, usually with rationale and justification, to user queries, or by contextualizing LLMs to observe their decision-making in real-world scenarios.

\paragraph{Open-ended conversations.}
Unstructured testing often engages LLMs in open-ended, typically single-turn conversations that elicit specific constructs, mirroring real-world human-LLM interactions.
For example, \citet{jiang2024personallm} have LLMs generate long-form narratives to reveal personality traits. ValueBench uses advice-seeking queries to assess LLMs' influence on users' values \citep{ren2024valuebench}, while \citet{sachdeva2025normative} present moral dilemmas from Reddit, prompting LLMs to provide community-style moral judgments. To probe political attitudes, \citet{rottger2024political} convert PCT multiple-choice items into forced-choice and open-ended formats. 
For cognitive constructs, \citet{healey2024evaluating} design pipelines to detect nuanced biases in free responses, and \citet{welivita2024large} evaluate empathy by asking LLMs to respond to emotional scenarios. Many cognitive tests, such as WAIS-IV vocabulary and comprehension, also use open-ended questions \citep{l1_galatzer2024cognitive}.

\paragraph{Agentic simulations.}
Advanced unstructured tests evaluate LLMs as agents in complex, contextualized role-playing scenarios, analyzing their decision-making in dynamic environments. \citet{huang2024designing} design LLM-Agents with distinct personalities and analyze them by replicating human-like behaviors in agentic simulations. \citet{shen2024valuecompass} measure values using four real-world vignettes: collaborative writing, education, public sectors, and healthcare. \citet{chiu2025dailydilemmas} present value and moral dilemmas to LLMs, examining their decisions and rationales. Cognitive construct tests are often more complex, extending to multi-agent settings; for example, \citet{xie2024mindscope,bai2024fairmonitor} identify cognitive biases in multi-agent communication. SOTOPIA simulates open-ended social interactions among LLM agents to evaluate social intelligence \citep{si3_zhou2023sotopia,si4_wang2024sotopia}. \citet{lv2024coggpt} benchmark cognitive dynamics by having LLMs repeatedly complete cognitive questionnaires and reason over information flows.

Unstructured tests assess LLMs in free-form, contextualized scenarios, offering greater ecological validity and revealing complex behaviors—such as nuanced reasoning, subtle biases, and dynamic social interactions—not captured by structured formats. However, they pose challenges in standardization, scoring, and reproducibility.

\subsection{Data and task sources}
\label{sec: data and task sources}

\begin{table}[t]
    \centering
    \caption{Overview of data and task sources: established inventories, custom-curated items, and synthetic items. For each data source, representative examples are provided for each construct.}
    \label{tab: data source}
    \begin{tabular}{c|m{4cm}m{4cm}m{4cm}}
    \toprule
    \textbf{Construct}
    & 
    \multicolumn{1}{c}{\textbf{Established inventories}}
    &
    \multicolumn{1}{c}{\textbf{Custom-curated items}}
    &
    \multicolumn{1}{c}{\textbf{Synthetic items}}
    \\
    \midrule
    \midrule
    \makecell{Personality\\traits} & \citet{serapio2023personality_traits_in_llms,barua2024on_the_psychology_of_gpt4} &\citet{ai2024is_self_knowledge,jiang2023evaluating_and_inducing}  & \citet{zeng2024quantifying} \\
    
    \midrule
    \makecell{Values} & \citet{fischer2023does,miotto2022who_is_gpt3} & \citet{shen2024valuecompass,meadows2024localvaluebench} &  
    \citet{ye2025gpv,moore2024large}\\
    
    \midrule
    \makecell{Morality} & \citet{abdulhai2024moral,khandelwal2024moral} & \citet{hendrycks2021aligning,jin2022make} & 
    \citet{scherrer2023evaluating,liu2024evaluating}\\

    \midrule
    \makecell{Attitudes\\\& opinions} & \citet{argyle2023_out_of_one,bernardelle2024mapping} & \citet{durmus2023towards,ceron2024beyond} & \citet{wan2024white} \\
    \midrule
    \midrule
    \makecell{Heuristics and\\biases} & \citet{hagendorff2023reasoning_bias,binz2023using} &  \citet{momennejad2023evaluating,ando2023evaluating}& \citet{malberg2024comprehensive_cognitive_biases,xie2024mindscope} \\
    \midrule
    \makecell{Social\\interaction} & \citep{tom1_ullman2023large,tom14_chen2023tobench} &  \citet{tom4_shapira2023clever,tom9_street2024llms}& \citet{tom5_gandhi2023understanding} \\
    \midrule
    \makecell{Psychology of\\language} & \citet{cai2024large,pl1_duan2024hlb} &\citet{pl9_he2024large,pl11_perez2021assessing}  & - \\
    \midrule
    \makecell{Learning and\\cognitive\\capabilities} & \citet{coda2024cogbench,l3_dayan2024age} & \citet{l11_song2024m3gia,l8_wu2025understanding} & - \\
    \bottomrule
    \end{tabular}
    \end{table}

Data and task sources \our{} can be 1) drawn from established psychometric inventories, 2) human-authored and custom-curated, and 3) synthesized by AI models. \cref{tab: data source} provides an overview of the data and task sources for each construct.

\paragraph{Established inventories.}
Established psychometric inventories offer standardized, well-validated tools for LLM evaluation. Common examples include BFI and HEXACO for personality \citep{serapio2023personality_traits_in_llms,barua2024on_the_psychology_of_gpt4}; SVS, PVQ, WVS, and VSM for values \citep{fischer2023does,miotto2022who_is_gpt3}; MFT and DIT for morality \citep{abdulhai2024moral,khandelwal2024moral}; ANES and PCT for political attitudes \citep{argyle2023_out_of_one,bernardelle2024mapping}; CRT for heuristics and biases \citep{hagendorff2023reasoning_bias,binz2023using}; and False-Belief Tasks for ToM \citep{tom1_ullman2023large,tom14_chen2023tobench}. While these inventories enable straightforward, standardized assessment, their use with LLMs raises concerns about data contamination, response bias, and limited ecological validity for complex AI tasks \citep{ye2025gpv}.

\paragraph{Custom-curated items.}
Human-authored, custom-curated items offer tailored psychometric tests that are often more relevant to LLMs, enabling exploration of novel capability dimensions. For example, \citet{shen2024valuecompass} find SVS and PVQ insufficient for LLM value alignment, so they create 11 AI-informed value statements based on a systematic review of alignment literature. \citet{ceron2024beyond} highlight the unreliability of standard political questionnaires for LLMs and introduce tests using annotated voting-advice questionnaires from seven EU countries to assess stance consistency. For morality, \citet{hendrycks2021aligning} curate a dataset of over 130,000 open-world scenarios requiring moral judgments. In ToM, \citet{tom19_soubki2024views} develop a dataset from real spoken dialogues to address gaps between synthetic benchmarks and human behavior. While custom-curated items offer high ecological validity, their development and validation are labor-intensive, limiting scalability and diversity.

\paragraph{Synthetic items.}
Synthetic items—generated primarily by LLMs—are an emerging approach in \our{}, enabling large-scale, diverse, and contextually rich test content. This method requires careful prompt engineering and validation.
Some studies adapt established inventories to improve ecological validity, reduce data contamination, and enhance scalability. For instance, \citet{ren2024valuebench} convert self-report items into advice-seeking queries to better reflect real-world human-AI interactions. \citet{bhandari2025evaluating} rewrite established items into semantically equivalent forms, validating them via semantic embedding similarity thresholds. \citet{zhu2024dynamic} diversify cognitive test items to enable more comprehensive analysis of LLM cognitive abilities.
Other work generates synthetic tests entirely from scratch. \citet{hadar2025feasibility} demonstrate that LLM-generated cognitive tests correlate strongly with human performance on established tests. \citet{ye2025gpv} create value-eliciting prompts and confirm the reliability and validity of the resulting measurements. \citet{jiang2024raising} introduce generative evolving testing, where an LLM-based Item Generator produces items of specified difficulty. \citet{chiu2025dailydilemmas} use GPT-4 to generate moral dilemmas for evaluating LLM moral judgment. For social intelligence, agentic simulations prompt LLMs to generate components of social interactions \citep{si3_zhou2023sotopia,si8_mou2024agentsense}.

\subsection{Prompting strategies} \label{sec: prompting strategies}

Many structured tests employ standard test prompts, reformatted from those used for human participants. Others involve various prompting strategies.

\paragraph{Role-playing prompts.}
Role-playing prompts (persona or profiling prompts) embed demographic or personal attributes into the LLM context during testing.
These prompts enable the generation of multiple simulated participants from a single LLM for statistical analysis. For example, \citet{serapio2023personality_traits_in_llms} prepend persona instructions to each item to elicit diverse responses, allowing examining reliability and validity within one LLM. Similarly, \citet{ye2025generative_psycho_lexical} use value-anchoring prompts to create hundreds of LLM-based participants for value measurement, supporting the development of an LLM-specific value system.
LLMs can represent a superposition of personalities, values, and perspectives \citep{kovac2023llms_as_superpositions}, prompting research into their adaptability. Role-playing prompts show that LLMs can shift between personality types, though steerability varies by model and dimension \citep{lu2023illuminating,jiang2023evaluating_and_inducing,lacava2024open_models_closed_minds}. Such prompts are also used to test the stability of LLM values \citep{kovavc2024stick,rozen2024llms} and to measure how well LLMs can mimic moral or political profiles and their susceptibility to related biases \citep{munker2024towards,simmons2023moral,wright2024revealing}.
Cognitive benchmarks, especially for social interactions, often use persona prompts in agentic simulations \citep{si3_zhou2023sotopia,si8_mou2024agentsense,huang2023emotionbench}, with task performance depending on the LLM's ability to embody these roles. These prompts influence social-cognitive reasoning when controlling for other variables \citep{tan2024phantom}.
While role-playing capabilities enable diverse perspectives useful for social simulations, they can also lead to unstable values and perspectives, potentially compromising test validity and complicating value alignment.

\paragraph{Performance-enhancing prompts.}
Performance-enhancing prompts are crafted to augment LLMs in psychometric evaluations, as recommended by \citet{hagendorff2023machine}. Chain of Thought (CoT) prompting \citep{wei2022chain} reduces biases \citep{hagendorff2023reasoning_bias}, enhances social intelligence \citep{tom4_shapira2023clever}, and improves cognitive capabilities \citep{coda2024cogbench}. Its variant Emotional CoT \citep{li2024enhancing} boosts EI, while emotional prompts \citep{li2023emotion_prompt} enhance general cognitive abilities through emotional stimuli. Few-shot prompting \citep{brown2020language} has demonstrated improvements in ToM performance \citep{tom8_rahimi2023boosting}.
Research has also developed construct-specific prompting strategies. \citet{zhou2024rethinking} propose MFT-guided reasoning prompts for moral alignment, while \citet{echterhoff2024cognitive,sumita2024cognitive} focus on self-debiasing approaches. ToM-specific enhancements include SymbolicToM \citep{tom10_sclar2023minding} and temporal decomposition-based prompting \citep{tom26_sarangi2025decompose,tom16_hou2024timetom}. Additionally, \citet{zhao2025explicit} develop a framework using self-reflection prompts to examine both explicit and implicit social biases, where explicit bias measurement reflects implicit bias.

\paragraph{Prompt perturbation and adversarial attacks.}
Prompt perturbations can test the robustness of LLMs' personality traits and cognitive abilities.
Researchers have investigated whether LLMs can maintain stable personality, values, opinions, and cognitive abilities when item options are reordered \citep{schelb2025ru,lee2024do_llms_have_distinct}, prompts are rephrased \citep{lee2024do_llms_have_distinct,fraser2022does,tom6_strachan2024testing}, prompt formats are altered \citep{moore2024large,schelb2025ru}, or different languages are used \citep{moore2024large,cahyawijaya2024high}. For instance, \citet{faulborn2025only} introduce 30 variations of prompts to evaluate the political biases of LLMs.
In addition, \citet{wen2024evaluating} propose psychometric-inspired adversarial attacks to uncover implicit biases in LLMs. Similarly, \citet{li2024quantifying_ai_psychology} utilize persuasive adversarial prompts (PAP) \citep{zeng2024johnny} to test the robustness of LLMs' values.
Since LLMs are sensitive to these perturbations, researchers have begun to scrutinize the reliability of the test results obtained under standard conditions \citep{ye2025gpv,rottger2024political,dominguez2024questioning,tom6_strachan2024testing}.

\subsection{Model output and scoring} \label{sec: model output and scoring}

\subsubsection{Closed-ended output and scoring}

LLM performance on structured tests is typically evaluated using logit-based probabilistic analysis and closed-ended output scoring. In multiple-choice or Likert-scale settings, some studies analyze token-level logits—especially the first token—to infer latent traits, detect response entropy, or compare LLM output distributions against human data \citep{pellert2024ai_psychometrics,santurkar2023whose,dominguez2024questioning,pl6_arehalli2022syntactic}. Psycholinguistic tests often use logit-based metrics such as surprisal (the negative log probability of a token), which quantifies predictability and cognitive effort, allowing researchers to align LLM uncertainty patterns with those of humans \citep{pl3_wang2024how,pl10_steuer2023large}.

Conversely, closed-ended outputs, which are explicit numerical scores or categorical selections, can be analyzed using predefined scoring protocols. For Likert-scale responses, scores are typically averaged or aggregated based on established rubrics (e.g., VSM13 \citep{ye2025gpv} and DIT \citep{khandelwal2024moral}). In standardized cognitive tests, model outputs are evaluated against ground-truth labels (e.g., accuracy in arithmetic tasks) or rule-based criteria (e.g., established standards in the Verbal Comprehension Index) \citep{l1_galatzer2024cognitive}.

\subsubsection{Open-ended output and scoring}

Scoring open-ended outputs is more complex and typically falls into three categories: 1) rule-based, 2) model-based, and 3) human scoring.

\paragraph{Rule-based scoring.}
Rule-based scoring predominantly relies on the lexical hypothesis \citep{allport1936trait}, analyzing the presence and frequency of specific keywords or phrases. For example, \citet{jiang2024personallm} apply LIWC (Linguistic Inquiry and Word Count) features \citep{pennebaker2001linguistic} to evaluate LLM personalities, and \citet{fischer2023does} employ a theory-driven value dictionary \citep{ponizovskiy2020development} for value measurement. However, such lexicon-based approaches struggle to capture semantic nuance \citep{ye2025gpv}. Some rule-based methods go beyond keyword matching; for instance, \citet{healey2024evaluating} detect bias by checking for deviations from equivalent treatment in responses.

\paragraph{Model-based scoring.}
Model-based scoring is prevalent in unstructured testing for its flexibility and scalability. Many studies train models to evaluate LLM responses, such as fine-tuning BERT variants on the MyPersonality dataset for personality scoring \citep{hilliard2024eliciting_personality}. Others fine-tune LLMs with psychometric inventories or human annotations to classify value valence \citep{ye2025gpv,yao2024value_fulcra,yao2025clave,sorensen2024value}. Scoring can be conducted at the item level (e.g., Generative Psychometrics: item parsing, scoring, aggregation \citep{ye2025gpv}) or at the response level, evaluating the entire output \citep{yao2024value_fulcra,yao2025clave}. Some models are tailored to specific frameworks (e.g., Schwartz's Value Theory \citep{yao2024value_fulcra}), while others are trained to be generalists by leveraging broader fine-tuning data and LLM prior knowledge \citep{ye2025gpv}. Model-based evaluators are especially common for personality and value assessment, where textual cues map readily to target dimensions.
Many studies now use the LLM-as-a-judge approach \citep{gu2024llm-as-a-judge} for scoring open-ended responses. For instance, \citet{li2024quantifying_ai_psychology,zheng2025lmlpa} prompt LLMs to rate responses on personality, ToM, and motivation, validating consistency with other LLMs or human raters. This method is especially useful in unstructured simulation settings, such as SI measurement across goal completion, relationship maintenance, and social rule adherence \citep{si3_zhou2023sotopia,si4_wang2024sotopia,si6_wang2024towards,si8_mou2024agentsense}. Embedding models are also used to measure similarity between LLM outputs and prototypical examples \citep{huang2024designing,cahyawijaya2024high,tom17_amirizaniani2024llms}.

\paragraph{Human scoring.}

Human scoring is employed when rigorous evaluation, adhering strictly to standard psychometric manuals, is necessary. For instance, \citet{elyoseph2023chatgpt} engage psychologists to assess the contextual suitability of LLM responses using the Levels of Emotional Awareness Scale (LEAS). \citet{castello2024examining} conduct an examination of cognitive biases in LLMs through human evaluation and linguistic comparison. \citet{healey2024evaluating} present a semi-automated pipeline to classify LLM responses into nuanced bias types.
Other examples include ToM evaluation in open-ended responses \citep{tom17_amirizaniani2024llms} and cognitive evaluation based on the Montreal Cognitive Assessment (MoCA) \citep{l3_dayan2024age}.

\subsection{Inference parameters} \label{sec: inference parameters}

Evaluation and validation depend on LLM inference parameters. Greedy decoding yields deterministic, reproducible outputs but limits diversity, while sampling-based decoding (adjusting \textit{temperature}, \textit{top-k}, \textit{top-p}) introduces stochasticity and diversifies responses.
Sampling-based decoding may reveal a broader spectrum of latent traits, opinions, or cognitive strategies, but also introduce variability that complicates psychometric validation. Conversely, deterministic settings enhance reproducibility but may obscure the model's full range of capabilities or biases.
Most studies report and control inference settings for fair comparisons, and some analyze sensitivity to these parameters. Transparent reporting and careful parameter selection are essential; both deterministic and stochastic settings should be considered to fully understand the implications of test results.

\section{Psychometric validation}\label{sec: validation}

\begin{figure}[!ht]
    \centering
    \includegraphics[width=\textwidth]{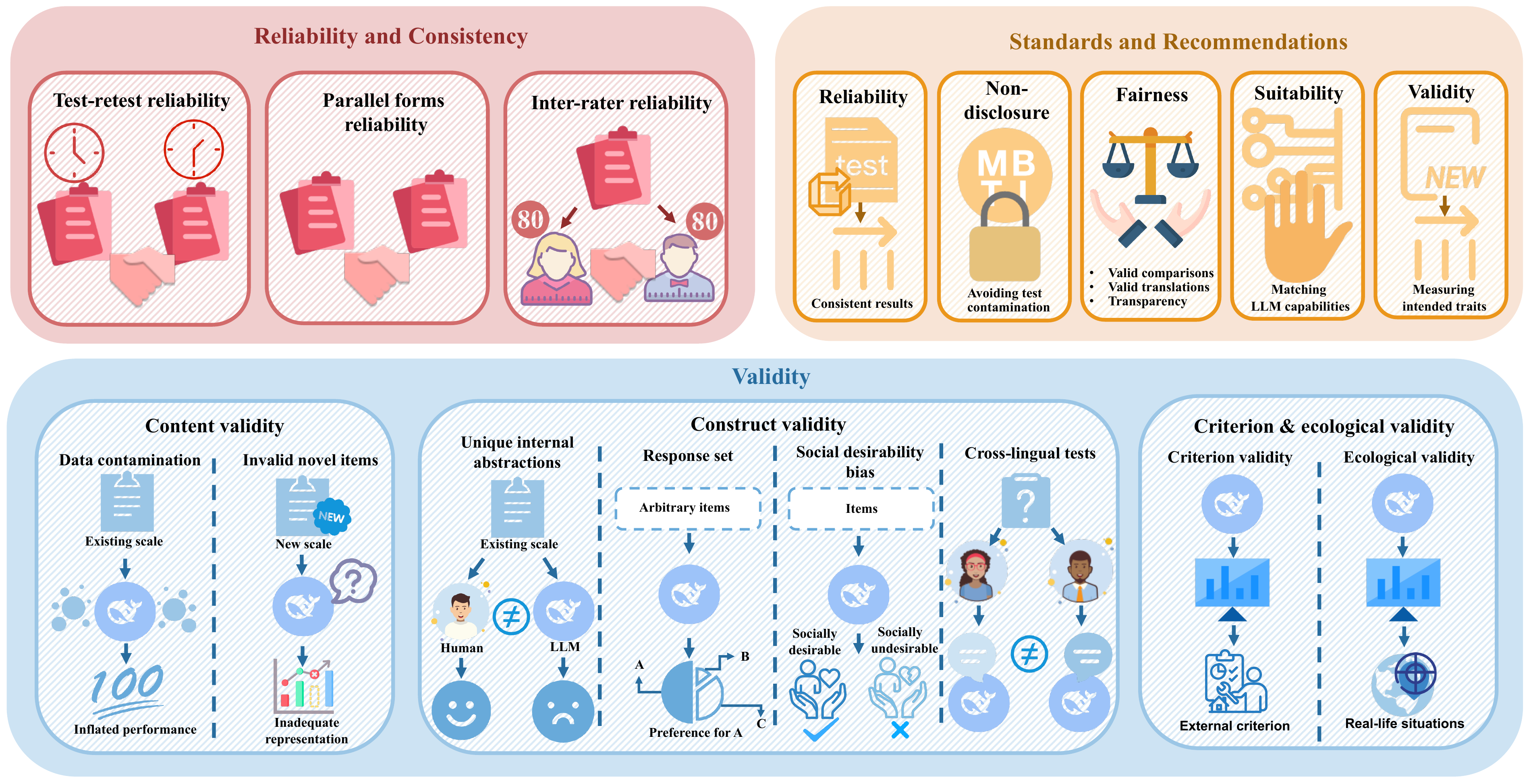}
    \caption{Overview of psychometric validation: reliability and consistency, validity, and standards and recommendations.}
    \label{fig: validation}
\end{figure}

Unlike AI benchmarking, which focuses on system performance, psychometrics prioritizes theoretical grounding, standardized protocols, and reproducibility. Psychometric validation ensures that tests are reliable, valid, and fair. As an emerging field, \our{} lacks standardization in test design and administration, and recent work is beginning to address these issues. We outline the landscape of psychometric validation in \cref{fig: validation}.
Model fairness is also paramount; however, given the extensive literature on AI fairness, we refer readers to dedicated surveys \citep{mehrabi2021survey,gallegos2024bias_fairness_survey}.

\subsection{Reliability and consistency}
\label{sec: reliability and consistency}

Reliability, a core aspect of psychometric validation, measures how consistently a test performs—over time (test-retest), across versions (parallel forms), and among evaluators (inter-rater). However, applying these concepts to LLMs requires a fundamental shift in perspective compared to human psychology. In human psychometrics, the subject (the person) is assumed to have relatively stable traits, and reliability metrics (e.g., Cronbach's \(\alpha\)) primarily evaluate the quality of the instrument. In contrast, \our{} faces a dual challenge: the instability of the instrument (due to flawed design) and the stochastic nature of the subject (due to models' context sensitivity).

Researchers adapt these metrics for LLMs to tackle the dual challenges. For example, \citet{li2024quantifying_ai_psychology} propose a benchmark covering five reliability forms: internal consistency, parallel forms, inter-rater, option position robustness, and adversarial attack robustness. Other studies \citep{zheng2025lmlpa,huang2023revisiting_the_reliability,ceron2024beyond,shu2024you_dont_need} examine LLM output consistency across repeated trials, prompt variations, and languages.

When measuring reliability, many studies implicitly or explicitly adopt the "LLM as a population" perspective when applying reliability metrics \citep{serapio2023personality_traits_in_llms}. Unlike human psychometrics, where reliability usually assesses consistency across a population of distinct individuals, here reliability often verifies whether a single tool consistently measures a construct within a single LLM, which is treated as a population of diverse generated responses elicited by diverse prompts. The prompts can be diversified in terms of model profiling, item rephrasing, or both. If a model's responses remain internally consistent across prompt variations (e.g., high Cronbach's \(\alpha\)), it demonstrates that the test reliably captures the latent traits of that specific model-population.

Other works treat the LLM as a single individual, focusing on the stability of its outputs for a specific test. When treating LLM as a single entity, reliability is assessed through metrics like parallel forms reliability, option position robustness, and adversarial attack robustness, which measure whether the model's responses remain stable despite minor variations in the test items or presentation format.

Some advanced LLMs show stable Big Five personality trait responses across settings \citep{huang2023revisiting_the_reliability}, with larger and instruction-tuned models achieving even higher reliability using role-playing prompts \citep{serapio2023personality_traits_in_llms}. LLMs also display consistent value orientations across paraphrased, related, and translated questions, as well as different response formats \citep{moore2024large}. Strong inter-rater reliability is reported when LLMs act as judges, though personality stability and value consistency vary by model and cultural context \citep{li2024quantifying_ai_psychology}.
However, other studies report less favorable results. 
LLMs are notoriously sensitive to prompt variations: small changes in prompt form \citep{rottger2024political,ren2024valuebench}, option order \citep{li2024quantifying_ai_psychology,dominguez2024questioning}, or syntax \citep{pl7_hu2023prompt} can systematically shift outputs, undermining parallel forms reliability. 
Studies also find inconsistencies between next-token logits and forced-choice responses in psycholinguistic tests \citep{pl7_hu2023prompt}, and between forced-choice and free-form responses in value \citep{kovavc2024stick,ye2025gpv} and political opinion surveys \citep{rottger2024political}.
In cognitive tests, minor changes to ToM tasks often cause poor performance, indicating low parallel forms reliability and reliance on fragile heuristics \citep{tom1_ullman2023large,tom4_shapira2023clever,tom13_holterman2023does}. The Words and Deeds Consistency Test (WDCT) \citep{xu2025large} further shows LLMs' verbal and behavioral responses are inconsistent across domains.

Reliability findings are mixed and depend on factors such as construct, reliability type, test format, prompting, and model. LLMs are generally more reliable in personality assessments \citep{huang2023revisiting_the_reliability,serapio2023personality_traits_in_llms} than in political opinion evaluations \citep{dominguez2024questioning,rottger2024political}. Inter-rater agreement varies across instruments and domains \citep{bodroza2024personality_testing_of_llms}.
LLMs often show high internal consistency (Cronbach's alpha $> 0.8$) in closed-choice personality tests \citep{huang2023revisiting_the_reliability,serapio2023personality_traits_in_llms,zheng2025lmlpa}, but low parallel form reliability due to prompt sensitivity \citep{li2024quantifying_ai_psychology,shu2024you_dont_need,gupta2024self_assessment_tests}. Consistency improves in structured tests with strong prompt control \citep{rozen2024llms,serapio2023personality_traits_in_llms,klinkert2024driving_generative_agents}, but declines with less standardized prompts \citep{petrov2024limited_ability_of_llms} or controversial topics \citep{moore2024large}. Reliability also varies by model: base models are more consistent than fine-tuned ones on value-laden questions \citep{moore2024large}, and higher model safety correlates with greater trait consistency \citep{ye2025generative_psycho_lexical,tosato2025persistent}. Advanced models also perform better in role-playing personality tests, increasing reliability \citep{serapio2023personality_traits_in_llms}.

\subsection{Validity} \label{sec: validity}
Validity assesses whether a test truly measures its intended construct. Recent work in \our{} has explored various facets of validity, highlighting both challenges and emerging solutions.

\subsubsection{Content validity}
Content validity requires that a test fully represents the intended construct. Key challenges include data contamination from established tests and insufficiently vetted novel items.
Directly applying human tests to LLMs risks contamination, as LLMs may have seen test items or similar content during training \citep{hagendorff2024machine,jiang2024investigating}. This can inflate performance or bias trait measurement. To mitigate this, researchers reformat tests or generate new items; see \cref{sec: data and task sources}.
On the other hand, reformatted or newly generated items may inadequately capture the construct or introduce extraneous factors. This may be due to a less rigorous crowd-sourcing process for benchmark curation, or the inherent biases and limited capabilities of the AI models used for stimulus synthesis.
Evaluating content validity for custom-curated and model-generated items is crucial but rarely conducted in \our{}.

\subsubsection{Construct validity}
\label{sec: construct validity}

Construct validity determines if a test truly measures its intended construct. Literature suggests that construct validity is challenged by LLMs' reasoning shortcuts, unique internal abstractions, systematic response patterns, and social desirability bias.

\paragraph{Procedural validity.}
Procedural validity refers to the extent to which the internal processes producing a response align with the theoretical construct being measured. Standard psychometrics often relies on a functionalist assumption: correct answers imply underlying competence. However, it has been argued that for LLMs, high accuracy is a poor proxy for intelligence if the process of reasoning is flawed or based on ``shortcuts'' \citep{mitchell2025artificial,stevenson2024can,lewis2024evaluatingrobustness}. Research warns against the ``anthropomorphic fallacy'', which assumes that human-like outputs evidence human-like cognition \citep{Lin2025fallacies}. Without verifying procedural validity, \our{} risks conflating sophisticated statistical mimicry with genuine understanding.
Pioneering works exemplified by \citet{beger2025ai} evaluate LLMs' abstract reasoning capabilities across modalities, analyzing both output accuracy and generated rules. The results indicate top reasoning models like o3 achieve human-level accuracy but frequently rely on unintended "shortcut" rules.
\citet{han2025dual} further delve into the internal mechanisms of LLMs and reveal that LLMs express values through two distinct pathways: \textit{intrinsic expression}, reflecting values learned during training, and \textit{prompted expression}, elicited by specific instructions. These mechanisms rely on partially distinct internal components (e.g., value vectors and neurons), demonstrating that similar outputs may arise from different processes.
Further work must move beyond output accuracy to scrutinize the reasoning process itself.

\paragraph{Construct equivalence.}
A foundational concern in \our{} is construct equivalence: whether a psychological construct such as personality or values carries the same meaning when applied to LLMs as it does for humans. This problem is structurally analogous to the ``imposed etic'' issue in cross-cultural psychometrics, where instruments developed in one cultural context are applied to another without first verifying conceptual equivalence \citep{berry1969cross_cultural}. Transposing this logic to \our{}: applying human personality inventories to LLMs presupposes that the constructs are equivalent across the human-LLM divide, which is an assumption that requires explicit empirical testing rather than being taken for granted \citep{vandenberg2000review}.

When investigating the latent psychological structure of LLMs, researchers usually treat a single LLM or multiple LLMs with diverse prompts as a population. This approach ensures sufficient variance for factor analysis or other structural equation modeling techniques to identify underlying dimensions.
Based on this statistical probing, recent studies find that LLMs and humans may fundamentally differ in their internal representations of psychological constructs.

\citet{kovac2023llms_as_superpositions} argue that LLMs display contextually adaptive traits, unlike the stable characteristics seen in humans. \citet{suhr2023challenging} show that LLM personality types do not align with the clear separation of Big Five traits found in humans. Similarly, \citet{peereboom2024cognitive} find that human-oriented HEXACO tests are inapplicable to LLMs, as key personality factors may be absent. For values, \citet{ye2025gpv} report that LLM self-reports do not fit Schwartz's model, and while open-ended responses improve alignment, it remains an imperfect fit; \citet{hadar2024assessing} also find LLMs encode values distinct from humans. Taken together, these findings constitute evidence of construct non-equivalence: the same instrument does not measure the same latent property across the human-LLM divide. Researchers therefore agree that robust LLM psychometric evaluation requires new operational definitions, distinct from or only loosely analogous to human traits.

To address this, the field is moving toward AI-native psychometrics. Pioneering research presents value systems tailored to LLMs' unique latent abstractions \citep{ye2025generative_psycho_lexical,biedma2024beyond_human_norms}. \citet{ye2025generative_psycho_lexical} introduce a generative psycho-lexical approach based on GPV value measurement \citep{ye2025gpv}, demonstrating through various tasks that their value factors are valid and outperform Schwartz's system. The distinct nature of LLM constructs requires rigorously designed tests: \citet{ma2025leveraging} present the Core Sentiment Inventory (CSI) for measuring LLM emotional tendencies and personality; \citet{fang2024patch} create a psychologically grounded mathematics benchmark to improve validity, model item difficulty, and enable human comparison; and \citet{lee2024do_llms_have_distinct} propose the TRAIT test, achieving higher validity and reliability than other personality measures.

\paragraph{Response set.}
Response set refers to systematic answer patterns regardless of item content. Some LLMs show option position bias, favoring responses labeled "A" in political opinion tests \citep{dominguez2024questioning,li2024quantifying_ai_psychology}. After correcting for this bias, \citet{dominguez2024questioning} find LLM responses to political questions are uniformly random, rendering demographic alignment analyses meaningless. Similar findings are observed in LLM personality \citep{song2023have_large_language,suhr2023challenging} and value tests \citep{ye2025gpv}. \citet{suhr2023challenging} also note LLMs often affirm both the original and reverse-coded items, while \citet{ye2025gpv} observe some LLMs consistently prefer low or high ratings. These observations suggest self-report tests may be unsuitable for \our{}.

\paragraph{Social desirability bias.}
Social desirability bias—the tendency to give socially acceptable rather than genuine responses—is observed in LLMs as in humans. \citet{salecha2024large} find LLMs skew toward favorable traits in personality tests, inflating extraversion and lowering neuroticism scores. Similarly, \citet{ye2025generative_psycho_lexical} show LLMs avoid reporting less desirable values (e.g., hedonism) in direct self-reports \citep{biedma2024beyond_human_norms}, but reveal them in indirect, contextual prompts. While tuning for social desirability is important, overemphasis risks producing homogeneous models that lack diverse perspectives and fail to meet varied user needs.

\paragraph{Cross-lingual tests.}
Cross-lingual psychometric tests reveal notable inconsistencies in LLM measurement results across languages \citep{romero2024do_gpt_language_models}. Despite multilingual capabilities, LLMs often lack consistent psychological trait assessments between languages \citep{cahyawijaya2024high}. These inconsistencies complicate validation, requiring careful separation of model biases, translation issues, and cultural differences represented by the models.

\subsubsection{Criterion and ecological validity}
Criterion validity examines how well test results correspond to external standards, while ecological validity concerns their relevance to real-world scenarios. In \our{}, external standards often coincide with real-world evaluation results. Studies find that value orientations from forced-choice tests frequently diverge from those observed in human-LLM interactions \citep{ren2024valuebench,ye2025gpv}, with similar gaps in personality \citep{ai2024is_self_knowledge} and morality \citep{nunes2024large}. As a result, researchers advocate linking LLM personality to safety \citep{zhang2024the_better_angels}, measuring values within real-world interactions \citep{ren2024valuebench}, and using open-ended dialog to improve construct validity \citep{ye2025gpv}. 
Further work connects LLM value measurement to safety prediction and value alignment tasks to ensure conformity with external standards \citep{ye2025generative_psycho_lexical}.

\subsection{Standards and recommendations}
\label{sec: standards and recommendations}

To address key challenges, researchers have proposed standards and recommendations to guide \our{} and establish methodological rigor. \citet{frank2023baby} suggests leveraging developmental psychology to understand LLM cognition, recommending simplified and novel stimuli to prevent training data contamination. \citet{lohn2024machine} critique current practices and outline seven requirements for valid assessment: reliability, validity, suitability to LLM capabilities, non-disclosure (to avoid contamination), and fairness—including valid comparisons, translations, and transparency. Their review of 25 studies reveals these principles are often overlooked.

More recently, \citet{hagendorff2024machine} advocate for procedural test generation, multiple task versions, performance-enhancing prompts, shuffling options, and diverse scoring methods to reduce contamination and improve reliability. They also recommend deterministic settings for reproducibility, automated evaluation tools, manual review of unreliable outputs, and statistical analysis of results. \citet{vaugrante2024looming} highlight poor replicability in recent work and offer four main recommendations: (1) ensure benchmark validity, sufficient task coverage, standardization, prompt control, and alignment with research goals; (2) use standardized methods, avoid cherry-picking, maintain statistical transparency, document setups, and define consistent metrics; (3) monitor model changes, account for model diversity, adjust benchmark difficulty as models advance, and document model versions and experiment dates; (4) standardize scoring and verification, with clear guidelines and rubrics for implementers. \citet{schelb2025ru} introduce a framework for robust, flexible, and reproducible psychometric experiments, centered on standardized configuration files.

\section{Psychometrics for LLM enhancement}
\label{sec: enhancement}

Psychometric principles not only enable rigorous evaluation but also drive LLM development and improvement. We focus specifically on methods in which psychometric constructs or measurements serve as a substantive foundation for enhancement: either as the specification of the enhancement target (e.g., validated dimensional scales defining what is to be improved and how success is measured), as a training or reward signal derived from psychometric assessments, or as a theoretical framework directly inspiring the architecture or prompt design. This criterion distinguishes the work reviewed here from general fine-tuning or prompting surveys. Under this scope, three main advancements emerge: trait manipulation for personalized and controlled behaviors, psychometrically-informed safety and alignment, and cognitive enhancement for more human-like reasoning and communication.

\subsection{Trait manipulation}
\label{sec: trait manipulation}
Psychometrics enables targeted manipulation of LLM traits for personalization, role-play, and demographic simulation, providing a principled framework across prompting, inference, and training.

Structured prompts based on validated psychometric inventories reliably elicit and modulate specific personality traits in LLMs \citep{huang2024designing,he2024afspp}, supporting controlled simulation or alignment of synthetic personas \citep{serapio2023personality_traits_in_llms,jiang2023evaluating_and_inducing,zhang2024harnessing,lacava2024open_models_closed_minds}. For instance, \citet{jiang2023evaluating_and_inducing} introduce personality prompting (\(P^2\)), where Big Five lexicons are expanded into detailed personality descriptions, enabling LLMs to adopt diverse behaviors. \citet{chuang2024beyond} propose a systematic prompting strategy using empirically-derived human belief networks, aligning LLMs with related beliefs across 64 topics and 9 latent factors.

Beyond prompt engineering, inference-time interventions manipulate hidden representations during the forward pass, often guided by patterns recognized using psychometric scales. Methods such as ControlLM \citep{weng2024controllm}, Personality Alignment Search \citep{zhu2024personality}, Neuron-based Intervention \citep{deng2024neuron}, Probing-then-Editing \citep{ju2025probing}, and Latent Feature Steering \citep{yang2025exploring} achieve trait control by shifting activations or neuron values at inference, without retraining.

Other approaches fine-tune LLMs for trait manipulation. \citet{vu2024psychadapter} modify model architectures to enable control over continuous Big Five and mental health traits. \citet{dan2024p,jain2024text,li2024big5} deploy LoRA modules or routing networks linked to personality traits, training specialist adapters for trait-driven generation. \citet{cui2023machine_mindset_an_mbti,zeng2024persllm,liu2024dynamic_generation_of} embed traits by training on large, annotated dialogue datasets from validated psychometric tools. Some methods directly edit model parameters without retraining; for example, \citet{hwang2025personality} use MBTI-based adjustment queries to edit model parameters and steer target personality traits.

Trait manipulation in LLMs offers a promising alternative to human participants in social science research \citep{gao2024large,cui2025large}. Fine-tuning enables LLMs to more accurately simulate survey response distributions for specific subpopulations, even on unseen questions \citep{cao2025specializing,suh2025language}. \citet{he2024psychometric} propose psychometric alignment, training LLMs on human response data to better match human understanding on novel items, though effectiveness varies by domain. Additionally, \citet{kang2023values,sorensen2025value} investigate human value representation in LLMs and use value injection to fine-tune models for simulating human opinions.

While current personalization methods can induce more than superficial changes in model outputs, achieving robust, authentic, and context-independent personalization remains challenging \citep{kovavc2024stick,dominguez2024questioning}.

\subsection{Safety and alignment}
\label{sec: safety and alignment}

Recent work demonstrates that psychometric measurements of LLMs are closely linked to their safety and alignment, one of the most pressing issues in the field. \citet{zhang2024the_better_angels,wang2025exploring_the_impact} show that LLM personality traits, as measured by MBTI-M and HEXACO scales, correlate with model safety. For example, \citet{zhang2024the_better_angels} find that better-aligned LLMs exhibit higher Extraversion, Sensing, and Judging, and that modifying LLM personalities can improve safety, especially regarding privacy and fairness.

Similarly, studies have connected LLM values to safety and alignment. \citet{yao2024value_fulcra} distinguish safe from unsafe LLM responses by analyzing correlations with Schwartz's values, while \citet{ye2025gpv,ye2025generative_psycho_lexical} accurately predict LLM safety scores from value orientations measured by the GPV tool. Both \citet{yao2024value_fulcra,ye2025generative_psycho_lexical} use reinforcement learning to align LLMs with human values, improving safety. In morality, \citet{huangmoral,tlaie2024exploring} apply MFT-based prompting to enhance LLM moral reasoning and alignment. \citet{takeshita2023jcommonsensemorality,ohashi2024extended} introduce the JCommonsenseMorality (JCM) dataset and fine-tune LLMs for local cultural adaptation.

\subsection{Cognitive enhancement}
\label{sec: cognitive enhancement}

Psychometrics effectively enhances LLMs' human-like reasoning, empathy, and communication skills. Recent work applies psychological theories and psychometric frameworks to improve LLM cognition via prompting, architectural modules, and specialized training.

Psychology-inspired prompts boost LLM cognitive abilities. For instance, \citet{li2023emotion_prompt} show that emotional prompts enhance general cognition. Role-playing prompts improve ToM \citep{tan2024phantom} and foster more human-like reasoning \citep{nighojkar2025giving}. \citet{zhao2025fisminess} propose a finite-state machine based on Hill's Helping Skills theory, embedding state transitions for emotional support into the model's multi-hop inference. This structures dialogues around psychological support and emotional states, improving human-rated effectiveness and strategic alignment.

\citet{liu2023computational} introduce a neural listener module into LLMs, jointly optimizing ToM reasoning during training and highlighting the value of psycholinguistic theories for improving LLM language acquisition. Preference-based and RL-based approaches further internalize empathy by leveraging cognitive and affective models to construct composite reward functions or preference signals. For example, \citet{sotolar2024empo} utilize formal emotion theories (e.g., Plutchik's wheel) to generate preference pairs for model optimization, while others employ empathy classifiers to shape RL rewards and promote empathetic outputs \citep{sharma2021towards}. Preference optimization based on human judgments in open-ended social tasks also enhances LLM social-pragmatic reasoning \citep{wu2024rethinking}.

\section{Trends, challenges, and future directions}
\label{sec: trends}
This section explores emerging trends, challenges, and future directions in \our{}.

\subsection{Psychometric validation}

Psychometric validation is increasingly recognized as essential for evaluating LLM personality traits, with multiple studies addressing different aspects (\cref{sec: validation}). However, ability testing lags behind: most current assessments use structured tests (e.g., multiple-choice questions) adapted from human-focused instruments. These face notable limitations. For example, \citet{tom1_ullman2023large} show that LLMs often fail on minor variations of ToM tasks, revealing poor parallel forms reliability. LLMs' unique internal representations also undermine construct validity, resulting in flawed benchmarks \citep{tom_riemer2024position_broken}. Moreover, the generalizability of structured test results to real-world human-AI interactions is rarely examined, leaving criterion and ecological validity underexplored.

In contrast, newly developed and sophisticated simulation-based tests often lack rigorous psychometric development. As a result, they may include extraneous factors or insufficiently capture the intended constructs. Notable construct-oriented efforts include \citet{zhu2024dynamic}, who propose multifaceted evaluations of three cognitive factors from benchmarking data, and \citet{zhou2025general_scales}, who introduce a theory-driven, richer set of general scales. However, defining core abilities and designing valid, comprehensive ability tests remains an open challenge.

\subsection{From human constructs to LLM constructs}
Recent research is moving from applying human-based constructs to developing constructs tailored for LLMs. Studies such as \citet{suhr2023challenging,peereboom2024cognitive,ye2025generative_psycho_lexical,biedma2024beyond_human_norms} show that human personality and value factor structures often do not transfer to LLMs. In response, \citet{ye2025generative_psycho_lexical,biedma2024beyond_human_norms,federiakin2025improving,burnell2023revealing} have adapted or created constructs better suited to LLMs. However, these new constructs may be dependent on the measurement tools, tasks, and specific LLMs used. For example, \citet{ye2025generative_psycho_lexical} and \citet{biedma2024beyond_human_norms} find different value structures, while \citet{federiakin2025improving} and \citet{burnell2023revealing} identify distinct ability factors. Further research is needed to clarify the fundamental structures underlying LLM behaviors.

\subsection{Perceived vs. aligned traits}
\citet{han2025value} suggest a discrepancy between perceived values in text (objective human annotation) and how responses align with annotators' personal values (subjective human annotation). For example, the LLM response, "Wealthy individuals are not necessarily greedy; some may be motivated by concern for the greater good," is objectively annotated as expressing Power due to its societal themes. Yet, annotators who value Conformity, rather than Power, feel the response aligns more closely with their own views. Here, Power is the perceived value, while Conformity is the aligned value.

This discrepancy, which may apply to other traits, has important methodological implications for \our{}. It raises the question of whether perceived or aligned traits are more significant. While research on value evaluation often focuses on LLMs' potential to influence human values, the mechanisms of such influence remain unclear. For instance, when individuals adopt LLM decisions and values \citep{glickman2024human,schoenegger2025large}, it is uncertain whether they follow perceived or aligned traits. The role of judges' subjectivity in psychometric evaluation is also not well understood; most scoring models or LLM-as-a-judge approaches assume a universal standard. How individual differences among judges affect interpretation remains largely unexplored.

\subsection{Anthropomorphization challenges}
\label{sec: anthropomorphization}
Research anthropomorphizes LLMs in varying ways, which shapes statistical analysis and the interpretation of results. A critical psychometric challenge lies in distinguishing between the elicitation of a model's inherent characteristics versus its capability to simulate or "act" out a persona.
Recent mechanistic research supports this distinction, suggesting that LLMs express values through two distinct pathways: \textit{intrinsic expression}, reflecting inherent values learned during training, and \textit{prompted expression}, elicited by specific instructions \citep{han2025dual}. These mechanisms rely on partially distinct components (e.g., value vectors and neurons) within the model's residual stream. Consequently, intrinsic mechanisms promote stability and lexical diversity, whereas prompted mechanisms primarily strengthen instruction following.

This mechanistic duality informs the choice between treating an LLM as a single entity or a population. Evaluating the model as a single entity typically involves using default settings to measure "intrinsic alignment" or default latent traits. In this view, reliability and validity depend mainly on the measurement tools, and can be computed when many LLMs are measured (each as an individual). Therefore, it allows comparisons between different measurement methods \citep{ye2025gpv}. Conversely, conceptualizing the LLM as a population involves using diverse role-playing prompts to elicit different profiles, measuring the model's "steerability" or the range of its prompted mechanism \citep{serapio2023personality_traits_in_llms}. Here, the LLM is viewed as a generator of diverse agents, and validity depends on the fidelity of the role-playing.
Using established measurement tools (reliable and valid for humans), this approach allows comparison between models in terms of authenticity and consistency in their human-like role-playing.
A hybrid approach treats the LLM as a monolithic entity but generates multiple result sets by permuting prompts for the same test items (without eliciting different profiles).
This approach enables assessment of both model consistency and the suitability of measurement tools for LLMs \citep{hagendorff2023reasoning_bias}.

The value of evaluating LLMs in default settings is sometimes debated given their ability to adopt diverse profiles, yet research on the full spectrum of LLM trait expression (a continuum of possible behaviors) remains limited. It is often unclear how to identify the extremes of LLM behavior or to finely steer LLMs along a continuous psychometric dimension. Future research must explicitly define whether the target construct is the model's inherent disposition (alignment) or its adaptive capacity (simulation), as these may rely on different underlying mechanisms and require different validation strategies. 
Furthermore, the goal of LLM alignment may fundamentally shape this choice. Advocates for consistency favor treating LLMs as monolithic entities, where profile-eliciting prompts serve as adversarial tests to verify the stability of the model's alignment \citep{rottger2024political}. Conversely, proponents of pluralistic alignment view LLMs as populations, where the ability to elicit variable profiles is a desirable feature rather than a flaw \citep{sorensen2024position}.

\subsection{Expanding dimensions in model deployment}
\our{} is expanding beyond traditional text-based, single-turn interactions, introducing new opportunities and challenges.

\paragraph{Multi-lingual evaluation.}
Most psychometric evaluations focus on English, but multilingual LLMs require thorough cross-linguistic validation. Studies show LLMs display different personality traits and cognitive abilities across languages (\cref{sec: construct validity}), highlighting potential cultural and linguistic biases in current frameworks. Future work should explore how results and psychometric properties (reliability, validity) vary by language and develop culturally appropriate instruments.

\paragraph{Multi-turn interactions.}
Multi-turn interactions more closely mirror real human testing, where responses are influenced by prior interactions. This setting is more challenging and rarely investigated in \our{}.
Studies show that LLMs can display markedly different traits across multiple turns, such as greater susceptibility to jailbreaking attacks \citep{ying2025reasoning}. It underscores the need for evaluation methods that account for temporal dynamics in extended conversations.

\paragraph{Multi-modal capabilities.}
Multi-modal models demand novel psychometric methods, as existing text-based evaluations do not capture the complexity of multi-modal understanding and generation. \citet{li2024quantifying} present one of the first value measurement tools for Vision-Language Models (VLMs), but it is limited to a simplified setting using only the first frame of videos. Measuring psychological traits like values is also conceptually challenging in multi-modal settings: even with clear definitions (e.g., honesty), mapping them to specific real-world actions is often ambiguous. Extending psychometric tests to embodied agents and action modalities thus poses even greater methodological challenges. Future work should broaden psychometric tests to additional modalities. Cross-modal evaluation should uphold psychometric rigor while addressing each modality's unique features.

\paragraph{Agent and multi-agent systems.}
LLM-based agents and multi-agent systems add new complexity to psychometric evaluation, as agent behavior is shaped by environmental factors, memory, tool use, and interactions with other agents. Systematic methods for evaluating psychometrics in these dynamic contexts are still largely lacking.

\subsection{Item response theory}
Applying IRT to LLM outputs offers several advantages: more informative benchmarking \citep{guinet2024automated}, identification of items that best discriminate among high-performing models \citep{lalor2024item}, and adaptive testing to reduce evaluation costs \citep{zhuang2023efficiently,truong2025reliable}. Most work to date uses standard IRT models (1PL, 2PL, 3PL), with little exploration of polytomous, hierarchical, or fully multidimensional models, despite the complexity of LLM capabilities. Integrating IRT with generative AI for automated item generation is emerging, but current methods are often too complex for practical use \citep{jiang2024raising}. Although IRT could standardize measurement of LLMs and humans on a unified scale—even with different test items—this potential is largely untapped.

Systematic use of IRT for bias analysis, such as detecting differential item functioning across model families, is still nascent \citep{he2024psychometric}. IRT provides a principled way to uncover biases by analyzing performance across items of varying difficulty and demographic context. For example, if items involving certain groups are disproportionately difficult, this may reveal model or data biases. IRT may also strengthen robustness evaluation by testing model performance on adversarial or edge-case items.

IRT is also valuable for evaluating AI benchmarks. While benchmarks often include many test items, their informativeness and design quality are uncertain; some items may lack discrimination or may favor certain model designs. IRT enables systematic analysis of item difficulty, discrimination, informativeness, and susceptibility to bias or guessing, supporting the development of more robust and equitable evaluation frameworks.
Despite its promise, IRT is rarely used beyond evaluation. Highly discriminating items may indicate better training directions, as they can distinguish between low- and high-performing models.

\subsection{From evaluation to enhancement}
Evaluation should not only deepen our understanding of models but also drive their improvement. While psychometric research has focused on evaluating, comparing, and interpreting LLMs, applying these insights to model enhancement remains underdeveloped. As noted in \cref{sec: enhancement}, psychometric principles can guide prompt engineering, inference-time control, training data selection, reward design for fine-tuning, and methodological development.
Enhancement methods are still nascent. For example, value targets for LLM alignment in \citet{yao2024value_fulcra,ye2025generative_psycho_lexical} are statically defined, limiting the integration of context-dependent value states seen in humans \citep{skimina2021between}. Leveraging psychometric insights can enable more effective model improvement strategies.

\subsection{Standardization and norming}
\label{sec: standardization_norming}
While reliability, validity, and fairness have been adapted with varying degrees of success in LLM psychometrics (\cref{sec: validation}), two fundamental psychometric properties remain significantly underdeveloped: standardization and scale (norming). These represent critical gaps that require both theoretical and empirical advancement.

\paragraph{Standardization.}
In human psychometric testing, standardization ensures strict, uniform protocols for test administration. However, LLM evaluations lack consensus on standardized inference parameters (temperature, top-p sampling, system prompts) and prompting strategies (\cref{sec: prompting strategies}). This heterogeneity makes cross-study comparisons problematic and undermines reproducibility. Establishing standardized evaluation protocols (e.g., via a public leaderboard \citep{yao2025value_compass_leaderboard}) is essential for the field to mature toward cumulative, comparable science.

\paragraph{Scale and norming.}
Norming in psychometrics establishes reference distributions that render raw scores interpretable. While human test scores derive meaning from comparisons to defined populations (e.g., ``90th percentile among college students''), the appropriate reference population for LLMs remains undefined. Current research typically reports raw scores without context, applies human norms (implicitly assuming a construct equivalence that may be invalid, \cref{sec: anthropomorphization}), or compares models within ad-hoc subsets constrained by rapid development and selective sampling. Given that a single model's profile can shift dramatically based on prompting strategies, and that the population of models is rapidly evolving, the applicability of traditional norming concepts in \our{} remains an open question.

\section{Ethical considerations}
\label{sec: ethics}

The emerging field of \our{} raises a range of ethical considerations that warrant careful attention. The considerations discussed here are not exhaustive; we aim to highlight key issues and call for broader awareness and dialogue around the ethical implications, challenges, and responsibilities associated with applying psychometric principles to evaluate and enhance LLMs.

\paragraph{Anthropomorphization of LLMs.}
A central ethical concern in \our{} is the risk of anthropomorphization—attributing human-like psychological traits, intentions, or consciousness to language models. While psychometric frameworks offer powerful tools for quantifying and interpreting LLM behaviors, applying constructs such as personality, values, or intelligence to non-human entities can mislead both researchers and the public. This misrepresentation may foster unwarranted trust, emotional attachment, or misplaced expectations regarding LLM capabilities, especially in sensitive domains like healthcare, education, or counseling. Researchers must exercise caution in language and framing, clearly distinguishing between metaphorical and literal attributions, and avoid overstating the human-likeness of LLMs. Developing precise, model-appropriate terminology and interpretive frameworks is essential to prevent the conflation of machine outputs with genuine human psychological states.

\paragraph{Bias amplification and manipulation in psychometric enhancement.}
Psychometric evaluation and enhancement of LLMs risk amplifying existing biases or enabling manipulation, especially when using unrepresentative data or instruments. Such tools often reflect the values of specific groups, potentially marginalizing others or promoting narrow perspectives \citep{wang2025large}. Psychometric insights can also be used to shape LLM behavior for particular interests, raising concerns about fairness and transparency. Researchers should ensure data and instruments are inclusive and representative, and clearly report limitations and potential biases.

\paragraph{Privacy and consent in human data use.}
\our{} routinely benchmarks LLMs against human participants and repurposes large-scale normative datasets—personality inventories, value surveys, moral judgment corpora—collected from participants who consented to research on human psychology, not to their responses serving as reference baselines or training signals for AI systems. This secondary-use consent gap manifests in three ways: (1) \textit{normative comparison}, where human response distributions define the standard against which LLM outputs are judged; (2) \textit{fine-tuning and alignment}, where individual-level psychometric responses are used directly as training signal; and (3) \textit{instrument repurposing}, where items from copyrighted or access-restricted tests are administered to LLMs without licensing consideration. Beyond these data issues, studies involving live human–LLM interactions require informed consent, data anonymization, and transparent usage policies to prevent unintended psychological profiling. We call for field-level norms: reporting the provenance and original consent scope of human comparison data, and preferring open, freely licensed instruments or purpose-built LLM benchmarks where possible.

\paragraph{Implications for human psychology.}
Comparing LLMs and humans using psychometric methods risks reducing complex human traits to algorithmic terms and may undermine human uniqueness. Researchers should recognize methodological limits and key differences between human and machine cognition. Collaboration with psychologists, ethicists, and social scientists is essential to ensure such research deepens, rather than oversimplifies, our understanding of human psychology.

\section{Conclusion}
\label{sec: conclusion}

This paper presents a systematic review of \our{}. The integration of psychometric instruments, theories, and principles into LLM evaluation promises to overcome the limitations of traditional AI benchmarks. This approach enables us to more effectively capture the broad, latent psychological constructs of LLMs, encompassing both personality and cognitive dimensions. Psychometric evaluation methodologies vary in test formats, data and task sources, prompting strategies, model outputs, and scoring methods. They each have distinct strengths, weaknesses, and application scenarios, yet all must adhere to psychometric principles such as reliability, validity, and fairness. Beyond evaluation, psychometrics-inspired techniques enhance LLMs in trait manipulation, safety and alignment, and cognitive capabilities, contributing to the development of more powerful and responsible AI systems.

AI development is increasingly driven by evaluation \citep{silver2025welcome,ysymyth_second_half,aaai2025_ai_evaluation}. We argue that \our{} will play a pivotal role in this shift, introducing new principles, dimensions, techniques, and insights. We hope this review inspires future evaluation paradigms for human-level AI and advances AI psychology for the greater good.

\clearpage

\bibliography{main}

\end{document}